\pdfoutput=1
\documentclass[11pt]{article}

\usepackage{nodalida2025}
\usepackage{hyperref}
\usepackage{url}

\usepackage{times}
\usepackage{latexsym}

\usepackage[T1]{fontenc}
\usepackage[utf8]{inputenc}

\usepackage{microtype}

\usepackage{inconsolata}

\usepackage{graphicx}
\usepackage{todonotes}
\usepackage{caption}
\usepackage{subcaption}
\usepackage{float}
\usepackage{booktabs}

\usepackage{tikz}
\newcommand{\ditto}{%
    \tikz{
        \draw [line width=0.12ex] (-0.2ex,0) -- +(0,0.8ex)
            (0.2ex,0) -- +(0,0.8ex);
        \draw [line width=0.08ex] (-0.6ex,0.4ex) -- +(-1.5em,0)
            (0.6ex,0.4ex) -- +(1.5em,0);
    }%
}

\aclfinalcopy

\title{Poro~34B and the Blessing of Multilinguality}

\author{
  \textbf{Risto Luukkonen\textsuperscript{1,2}}
  \hspace{0.5cm}\textbf{Jonathan Burdge\textsuperscript{2}}
  \hspace{0.5cm}\textbf{Elaine Zosa\textsuperscript{2}}
  \hspace{0.5cm}\textbf{Aarne Talman\textsuperscript{3}}
\\
  \textbf{Ville Komulainen\textsuperscript{1}}
  \hspace{0.5cm}\textbf{Väinö Hatanpää\textsuperscript{4}}
  \hspace{0.5cm}\textbf{Peter Sarlin\textsuperscript{2,5}}
  \hspace{0.5cm}\textbf{Sampo Pyysalo\textsuperscript{1}}
\\
\\
  \textsuperscript{1}TurkuNLP, University of Turku, Finland
  \hspace{0.5cm}\textsuperscript{2}Silo AI, Finland
\\
  \textsuperscript{3}University of Helsinki, Finland
  \hspace{0.5cm}\textsuperscript{4}CSC – IT Center for Science, Finland
   \hspace{0.5cm}
 \textsuperscript{5}Aalto University
\\
\tt{risto.m.luukkonen@utu.fi}\hspace{0.5cm}\tt{jonathan.burdge@silo.ai} \\
\tt{peter@silo.ai}\hspace{0.5cm}\tt{sampo.pyysalo@utu.fi}
}

\begin{document}
\maketitle
\begin{abstract}
The pretraining of state-of-the-art large language models now requires trillions of words of text, which is orders of magnitude more than available for the vast majority of languages. While including text in more than one language is an obvious way to acquire more pretraining data, multilinguality is often seen as a curse, and most model training efforts continue to focus near-exclusively on individual large languages. 
We believe that multilinguality can be a blessing: when the lack of training data is a constraint for effectively training larger models for a target language, augmenting the dataset with other languages can offer a way to improve over the capabilities of monolingual models for that language.
In this study, we introduce Poro~34B, a 34 billion parameter model trained for 1 trillion tokens of Finnish, English, and programming languages, and demonstrate that a multilingual training approach can produce a model that substantially advances over the capabilities of existing models for Finnish and excels in translation, while also achieving competitive performance in its class for English and programming languages. We release the model parameters, scripts, and data under open licenses at \url{https://huggingface.co/LumiOpen/Poro-34B}.
\end{abstract}

\section{Introduction}

Neural language models based on the transformer architecture \cite{vaswani2017attention} have led to substantial advances in natural language processing. Encoder-only transformer models such as BERT \cite{devlin2019bert} have advanced the state of the art in a broad range of classification tasks, while decoder-only models such as GPT \cite{radford2018improving} have redefined what can be achieved by generative models, opening new areas of study in prompting and in-context learning.
The success of these models is related in substantial part to their scaling properties: training larger models on more data leads to better results and even entirely new capabilities \citep{brown2020language}. Studies refining our understanding of the optimal balance of model size and training steps have increased the demands on data \citep{hoffmann2022empirical}, and many recent models optimize further for inference-time efficiency by training smaller models on more data \citep{sardana2023beyond}.

These developments have introduced increasing demands for textual data, with many recent models pretrained on a trillion tokens or more \citep[e.g.][]{touvron2023llama,almazrouei2023falcon,MosaicML2023Introducing,li2023starcoder,lozhkov2024starcoder2,groeneveld2024olmo}. While such resources can still be assembled from internet crawls for a few of the languages best represented online, for the vast majority of human languages we have already run out of data for training the largest of language models \citep{joshi2020state,villalobos2022will}. While it is standard to repeat training data, repetition can lead to reduced sample efficiency and degradation of performance \citep{hernandez2022scaling}: \newcite{muennighoff2024scaling} estimate that the value of repetition starts to diminish rapidly after four epochs and that repetition ceases to add information around 40 epochs. The availability of data is thus currently a limit for monolingual training for all but a few of the highest-resourced languages.

\begin{figure*}[t!]
\centering
\begin{subfigure}{0.4\textwidth}
  \centering
  \includegraphics[width=0.85\linewidth]{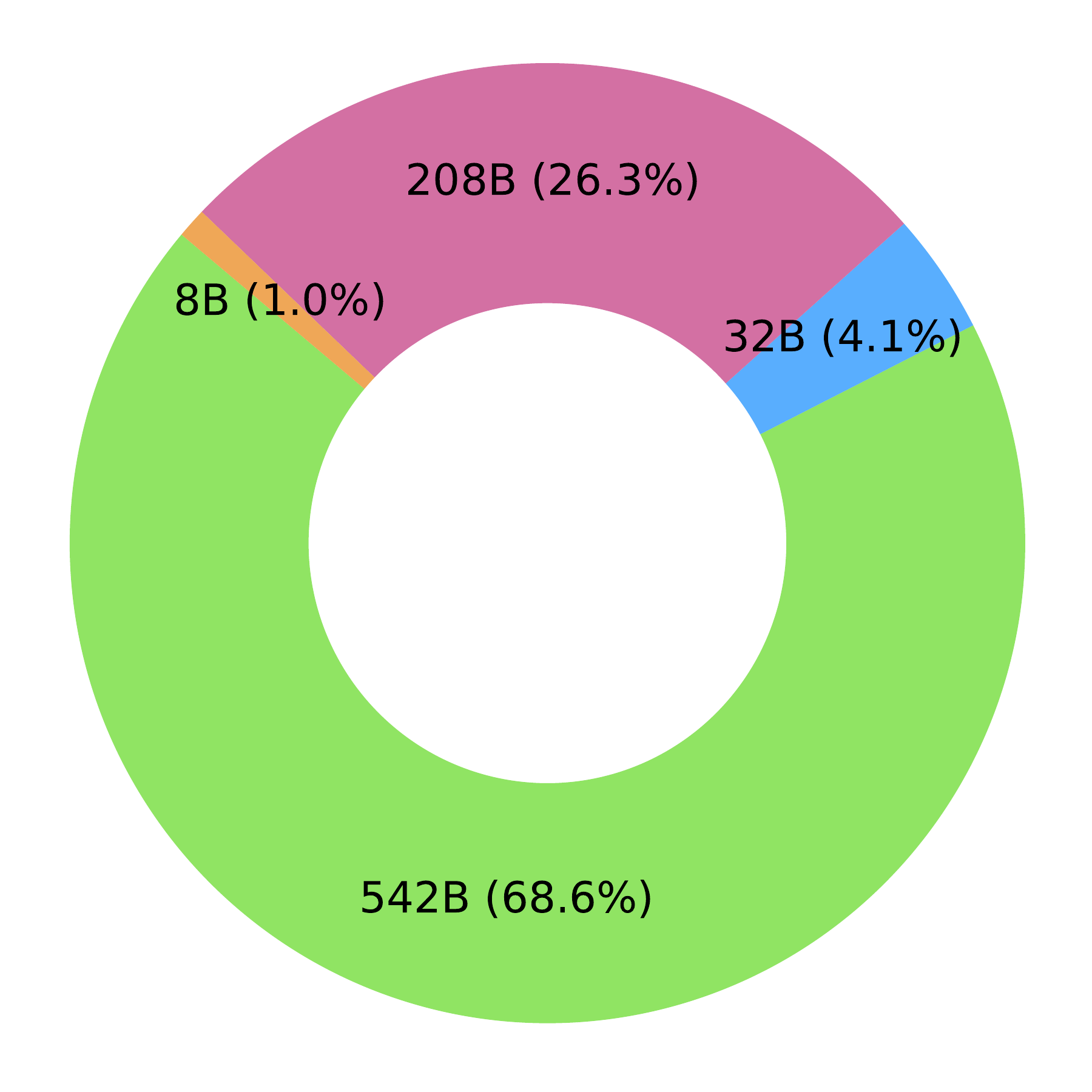}
  \caption{Original}
  \label{fig:orig-data-dist}
\end{subfigure}%
\begin{subfigure}{0.4\textwidth}
  \centering
  \includegraphics[width=0.85\linewidth]{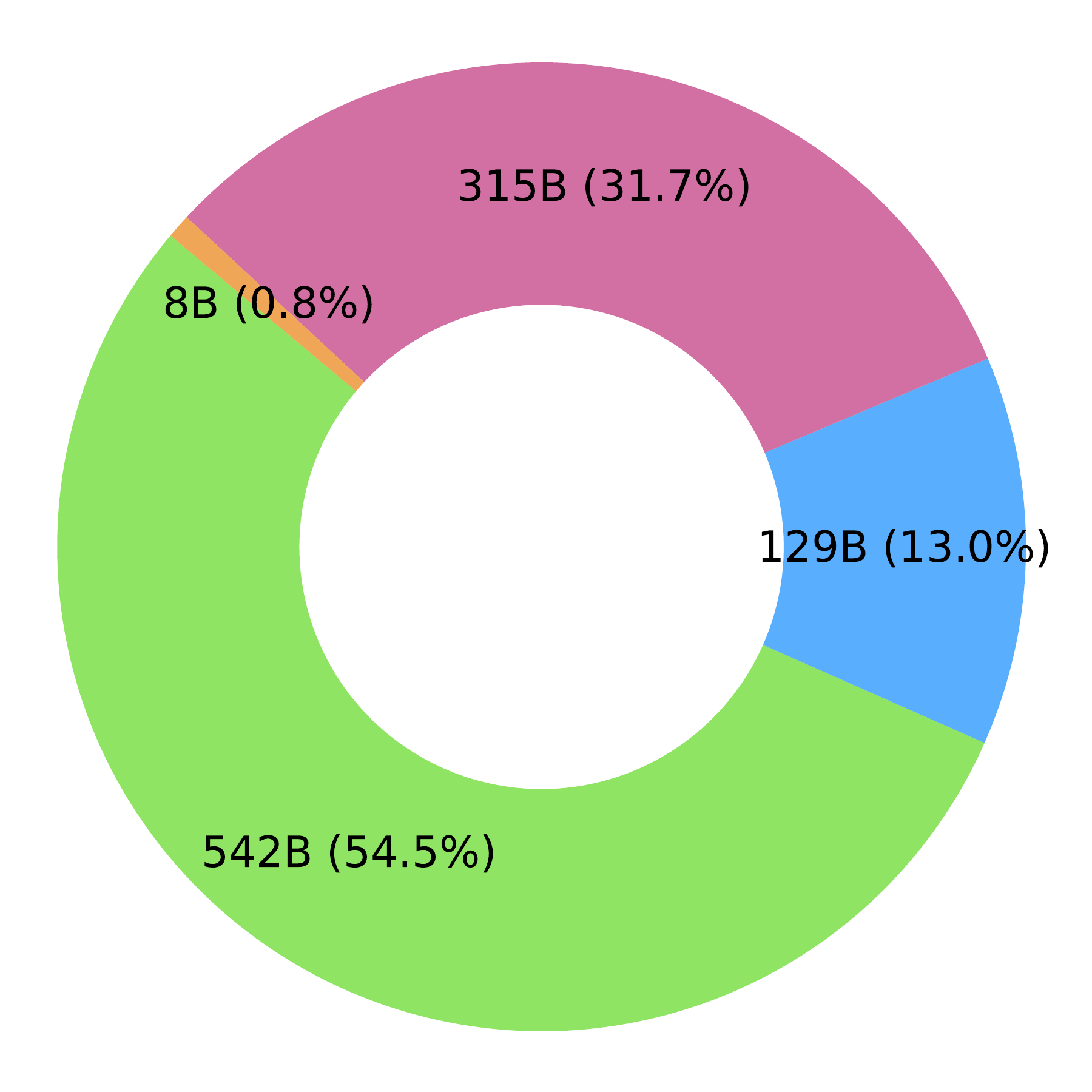}
  \caption{Sampled}
  \label{fig:sampled-data-dist}
\end{subfigure}%
\begin{subfigure}[t]{0.24\textwidth}
  \centering
  \includegraphics[width=0.45\linewidth]{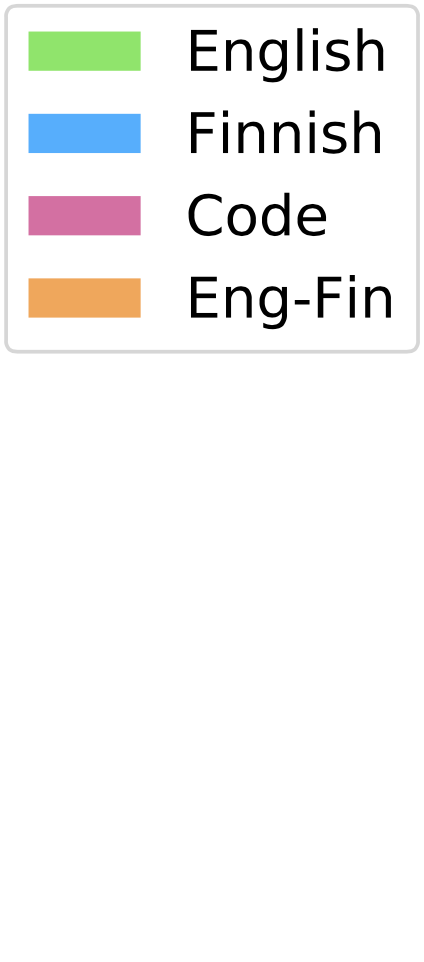}
  \label{fig:legend-data-dist}
\end{subfigure}
\caption{Pretraining data distribution.}
\vspace{-0.25cm} 
\label{fig:data-dist}
\end{figure*}

Multilingual training offers one obvious solution for increasing the amount of training data available, and a large number of multilingual transformer models have been introduced \citep[e.g.][]{conneau2020unsupervised,lin2022fewshot,lescao2022bloom,wei2023polylm}. However, despite the intuitive appeal of augmenting training data with texts in other natural languages, multilinguality is frequently seen as a negative -- commonly referred to as the \emph{curse of multilinguality} \citep{conneau2020unsupervised}. While there have been studies of the tradeoffs between monolingual and multilingual training \citep{fujinuma2022match,chang2023multilinguality} as well as efforts to enhance models specifically for multilinguality \citep{pfeiffer2022lifting} and to introduce additional language capabilities to existing models \citep{gogoulou2023continual,kew2023turning,zhao2024llama,ibrahim2024simple}, state-of-the-art generative models are still frequently trained near-exclusively on large languages such as English, with only limited efforts specifically focusing on optimizing performance for smaller languages.
In this study, we explore how to lift data limitations to create state-of-the-art large generative models from scratch for smaller languages, drawing on the understanding emerging in recent studies on how to make the most of limited data and assure that multilinguality is a blessing rather than a curse. Some key lessons from previous work include 
1) \textbf{limited multilinguality} instead of a large number of languages \citep{conneau2020unsupervised,chang2023multilinguality}
2) \textbf{matching scripts} (e.g., Latin) \citep{fujinuma2022match} and
3) \textbf{matching language families} \citep{pyysalo2021wikibert},
4) incorporating a \textbf{cross-lingual signal} using translation pairs \citep{anil2023palm,wei2023polylm},
5) \textbf{oversampling target language} data up to four epochs \citep{muennighoff2024scaling} and 6) augmenting natural language with \textbf{programming language data} \citep{madaan2022language,aryabumi2024code}.

We chose to specifically target the Finnish language, which is an interesting case for study as it is a Uralic language with no large close neighbours in its language family, necessitating more distant transfer than, for example, between English and another Germanic language. While the language is natively spoken by under six million people, its resources are still sufficient to consider a monolingual training approach for larger generative models. In a recent study, \newcite{luukkonen2023fingpt} combined several web crawls and curated sources of Finnish to create a dataset of approximately 40B tokens and introduced the monolingual FinGPT models trained from scratch for 300B tokens. With approximately 8 epochs, the repetition of data is expected to show diminishing returns \citep{muennighoff2024scaling},
and the largest of these models show signs of data limitations, with the 8B parameter model outperforming the 13B in benchmarks.
We believe it should be possible to overcome these limitations by applying the lessons listed above. While we cannot match language families, we train for four epochs over the Finnish data and augment it with both English and programming language data as well as an explicit cross-lingual signal from translation pairs. We pursue this approach to create Poro~34B, training a 34B parameter model for a total of 1T tokens -- 25 times more than the available Finnish data -- and evaluate the model in detail on Finnish, English, and programming language tasks. We find that the model not only achieves the goal of substantially advancing over the performance of existing Finnish models, but is also competitive in its class of open models on English and code as well as remarkably strong in translation tasks.

\section{Pretraining data}

For pretraining Poro~34B, we rely on datasets that have been previously preprocessed to remove low-quality texts and boilerplate,
filter toxic context, and deduplicate repeated texts.  We illustrate the pretraining data distribution in Figure~\ref{fig:data-dist} and describe the data briefly in the following. Data sources are detailed in Table~\ref{tab:data_sources} in the Appendix.

\paragraph{Finnish} For Finnish pretraining data, we draw on the resources recently introduced by \citeauthor{luukkonen2023fingpt} for creating the FinGPT model family. We exclude the \emph{ePub} and \emph{Lehdet} resources provided by the National Library of Finland for that work as they could not be shared due to copyright limitations, but use the remaining sources of data, totalling to a 32B token monolingual corpus. The majority of the Finnish data originates from web crawls (approx.\ 84\%) complemented with news sources (approx.\ 2\%), Project Lönnrot, the Finnish equivalent of Project Gutenberg copyright-free book corpus (approx.\ 0.5\%), Wikipedia (approx.\ 0.5\%) and Finnish online discussion forum contents from Reddit and Suomi24 (approx.\ 13\%). Following the rule of thumb proposed by \newcite{muennighoff2024scaling}, we upsample the 32B tokens of Finnish so that four epochs over the data are made during training. Consequently, approximately 13\% of the total tokens seen in pretraining are Finnish.

\paragraph{English} For English pretraining data, we primarily use SlimPajama~\citep{cerebras2023slimpajama}, a cleaned and deduplicated subset of the RedPajama corpus\footnote{\url{https://huggingface.co/datasets/togethercomputer/RedPajama-Data-1T}}~\citep{together2023redpajama}, from which we excluded data from the books category due to their copyright status.  We supplemented this dataset with the Project Gutenberg public domain books data from the Dolma corpus\footnote{\url{https://huggingface.co/datasets/allenai/dolma}}~\citep{dolma}. We train for one epoch over the 542B tokens of the English data, which thus represents slightly over half of the 1T total training tokens.

\paragraph{Programming Languages} To introduce data representing various programming languages (referred to hereinafter as ``code" for short) into our pretraining, we make use of the Starcoder corpus \citep{li2023starcoder}, a processed subset of The Stack corpus\footnote{\url{https://huggingface.co/datasets/bigcode/the-stack}} \citep{kocetkov2023stack}. The original corpus consists of 208B tokens, which we oversample 1.5x so that approximately a third of the tokens seen during pretraining represent code.

\paragraph{Cross-lingual data} We introduce a cross-lingual signal into pretraining by including translation examples from OPUS \citep{tiedemann2009news}.
Specifically, we use the English-Finnish examples from the 
Tatoeba dataset \citep{tiedemann2020tatoeba} to generate instruction-formatted translation examples.  The Tatoeba training data was reformatted into a minimalistic instruction-following format by recasting each English-Finnish translation pair into a document with the following format:

\begin{verbatim}
<|user|>Translate into Finnish: {{en}}
<|assistant|>{{fi}}
\end{verbatim}

\noindent
Where \texttt{\{\{en\}\}} and \texttt{\{\{fi\}\}} are the English and Finnish texts (resp.) of the translation pair. We additionally reverse the translation order (i.e., Finnish to English instead of English to Finnish) for a total of two documents for each sentence pair. No weighting is applied to the approximately 8B tokens of cross-lingual data, which thus represents slightly under 1\% of the pretraining tokens.

\section{Methods}

In this section, we describe the method used to create the Poro~34B tokenizer, the pretraining setup, and provide an estimate of the compute cost of pretraining the model.

\begin{figure*}[t!]
\centering
\includegraphics[width=0.95\textwidth]{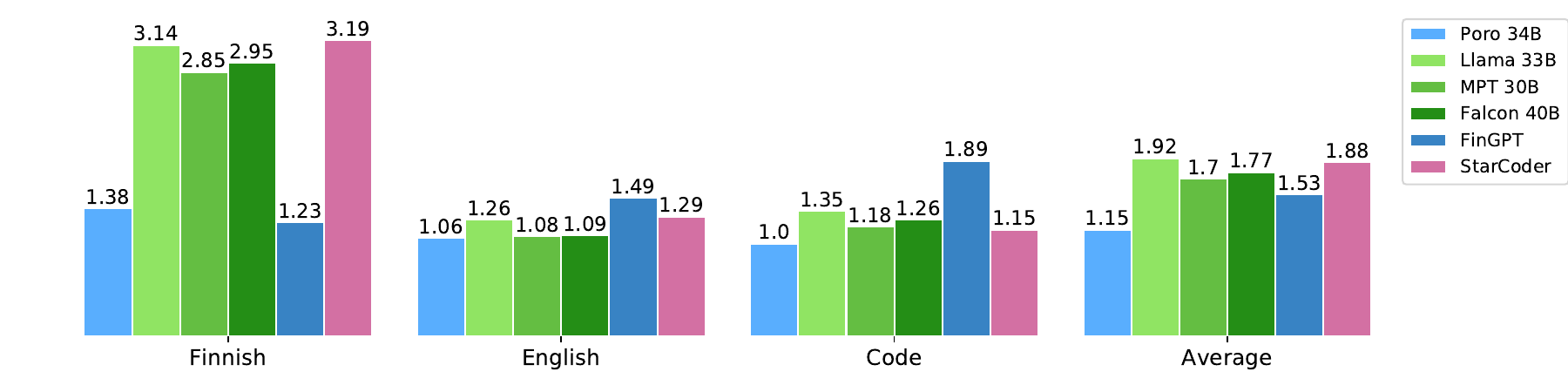}
\caption{Tokenizer fertility comparison (lower is better).}
\label{fig:fertility}
\end{figure*}

\subsection{Tokenization}
\label{sec:tokenization}

The choice of tokenizer has a broad range of impacts, not only on the efficiency of training and inference but also the capabilities of trained models \citep{rust2021good,petrov2023language,ali2023tokenizer}.
As we were not aware of any existing tokenizer that would be a good fit for our combination of languages and code, we created a new tokenizer for our model.
Specifically, we trained a custom byte-level BPE tokenizer using the same pre-normalization as the FinGPT tokenizer.
We selected a vocabulary size of 128K tokens, aiming to achieve low fertility on the targeted languages while keeping the vocabulary reasonably small. The tokenizer was trained on a uniform distribution of samples of the Finnish, English and code datasets. 

We assess the fertility of the tokenizer on the English and Finnish sentences from the devtest portion of the widely used Flores-101 benchmark for machine translation \cite{goyal2022flores}, which allows for a degree of cross-lingual comparability. For code, we use an approximately 1M character sample of lines from the Starcoder held-out test data.\footnote{We only sample lines with at least 10 alphabetic characters to avoid very short lines.} 
Figure~\ref{fig:fertility}
provides a comparison of the fertility of the tokenizer compared to selected reference tokenizers (see Section~\ref{sec:evaluation}). We find that on this data the new Poro~34B tokenizer has at least broadly comparable fertility to the lowest-scoring tokenizer on each of Finnish, English, and code, as well as the lowest average fertility of the compared tokenizers.

\subsection{Pretraining}

We next briefly present the key model and training parameters 
(detailed in Table~\ref{tab:model_arch_params} in the Appendix~\ref{sec:training_details})
and the pretraining software and configuration. 

\textbf{Architecture} Poro~34B is a decoder-only model with a parameter count of 34 billion, sharing its architecture with FinGPT  \citep{luukkonen2023fingpt} and BLOOM \citep{lescao2022bloom}. It incorporates layer normalization immediately following the input embedding layer for better training stability and uses ALiBi \citep{press2021train} as its positional encoding method. The model consists of 54 layers with a hidden dimension of 7168 and a total of 56 attention heads. 

\textbf{Training} We train to 1T tokens, intentionally exceeding the Chinchilla compute-opimality estimate \citep{hoffmann2022training} of approximately 700B tokens for a model of this size, thus gaining inference-time efficiency for the cost of additional compute investment in pretraining \citep{sardana2023beyond}. We train with a sequence length of 2048 tokens\footnote{We acknowledge that this can be considered limiting by today's standards, but this limitation can be relieved by methods for extending the context length, for example via linear extrapolation \citep{press2021train} or interpolation \citep{al2023position}.} 
using a cosine learning rate scheduler with a maximum learning rate of 1.5e-4, decaying to a minimum of 2e-5 over 990B tokens, and a linear warmup of 10B tokens. Our global batch size is 2048 samples totaling to 4194304 tokens per optimization step.

\textbf{Software} Poro~34B was trained on the LUMI supercomputer GPU partition, which is powered by AMD MI250X GPUs. The majority of open source frameworks for large language model pretraining are made to be primarily NVIDIA-compatible, and we required scalable AMD-compatible training software. Thus, we adopted the Megatron-DeepSpeed fork\footnote{\url{https://github.com/TurkuNLP/Megatron-DeepSpeed}} introduced by \cite{luukkonen2023fingpt},
which has optimized kernels converted from CUDA to be compatible with AMD ROCm, and has been demonstrated to be a viable solution for large model pretraining on LUMI. The hardware used to train the model is described in detail in Appendix~\ref{sec:appendix_hardware}.

\textbf{Configuration} Considering the hardware available and the selected hyperparameters such as batch size, a configuration of 128 nodes was chosen for the training of the model, resulting in a world size of 1024. The training was done using activation checkpointing, a micro batch size of 1, gradient accumulation of 16, and a 3D parallelism strategy of tensor parallel degree 2, pipeline parallel degree 4, resulting in a data parallel degree of 128. This allowed total training cycle throughput of 49618 TFLOPs and 174378 tokens/second.

\subsection{Compute cost}

Following \cite{groeneveld2024olmo}, we estimate the carbon footprint of our pretraining by multiplying the theoretical upper bound of the total power used by the GPUs when they are utilized at 100\% with the carbon intensity factor of LUMI. Taking into account the systems's power usage effectiveness (PUE) value of 1.04,\footnote{https://www.lumi-supercomputer.eu/sustainable-future/} we approximate the total power consumption to be 448MWh. As LUMI is powered by fully renewable electricity, we assume the carbon intensity factor to be 0.\footnote{We acknowledge that this assumption can be contested. As \cite{groeneveld2024olmo} note: "LUMI is powered entirely by
hydroelectric power and some sources \citep{ubierna2022water} measure the carbon intensity factor of hydroelectric power to be 0.024."} This brings our emissions to a total of 0 $\mathrm{tCO_{2}eq}$. It is important to note that we only take into account power consumption of the GPUs used, as the consumption of the entire node was not logged during training.

\begin{table*}[t!]
\centering
\setlength{\tabcolsep}{3pt}
\begin{tabular}{l|c|ccc|cc|c}
 & \textbf{Poro~34B} & \textbf{Llama 33B} & \textbf{MPT 30B} & \textbf{Falcon 40B} & \textbf{FinGPT 8B} & \textbf{FinGPT 13B} & \textbf{StarCoder} \\ \hline
Finnish & \textbf{1.89} & 2.98 & 2.89 & 3.57 & 1.94 & 1.92 & 3.83 \\
English & 1.87 & \textbf{1.81} & 1.89 & 1.85 & 2.55 & 2.46 & 2.38 \\
Code & 3.21 & 4.27 & 3.58 & 3.65 & 25.1 & 27.3 & \textbf{3.15} \\ \hline
Average & \textbf{2.32} & 3.02 & 2.79 & 3.02 & 9.86 & 10.6 & 3.12 \\
\end{tabular}
\caption{Character-level perplexity for Poro~34B and selected reference models (lower is better).}
\label{tab:char-ppl}
\end{table*}

\section{Evaluation}
\label{sec:evaluation}

We thoroughly analyze the capabilities of the model for Finnish, English and code, first briefly reporting perplexity results and then focusing on community-standard benchmarks for evaluating generative models. We then assess the quality of Finnish text generated by the model and finally evaluate the model's translation capability from English to Finnish (and vice versa). For comparison, we include results for the state-of-the-art Finnish language models, FinGPT~8B and FinGPT~13B~\citep{luukkonen2023fingpt}, and a selection of similarly-sized general-purpose open source base language models trained on broadly comparable numbers of tokens for English\footnote{We chose English models of similar size and training token budget rather than state-of-the-art models to more directly assess the effects of our multilingual training setup on performance in English.}: Llama~33B~\citep{touvron2023llama},
MPT~30B~\citep{MosaicML2023Introducing}, and Falcon~40B~\citep{almazrouei2023falcon}. We also provide results for
StarCoder~base~\citep{li2023starcoder} as a reference for performance on code tasks.

\subsection{Data and experimental setup} \label{data_and_experimental_setup}

We assess the perplexity of the model on the same data used to evaluate tokenizer fertility (Section~\ref{sec:tokenization}), namely Flores-101 devtest English and Finnish and a sample of the StarCoder test data.
As token-level perplexity is dependent on tokenization, it cannot be used to directly compare models with different tokenizers. We therefore report character-level perplexity $PPL_c$ following \newcite{ekgren2022lessons}, normalizing by character rather than token count when calculating perplexity.

We benchmark the capabilities of the model in Finnish using the FIN-bench\footnote{\url{https://github.com/TurkuNLP/FIN-bench}} dataset ~\citep{luukkonen2023fingpt}, which covers a variety of tasks to assess various aspects of model capabilities in Finnish, combining selected tasks translated and manually corrected from English BIG-bench \citep{srivastava2022beyond} with additional Finnish tasks. We evaluate all FIN-bench results in a 3-shot setting using the standard metrics defined for the benchmark. For English evaluations, we use LM Eval Harness~\citep{eval-harness} to evaluate with the following datasets:
ARC Challenge~\citep{Clark2018ThinkYH},
GSM8K~\citep{cobbe2021training},
HellaSwag~\citep{zellers2019hellaswag},
MMLU~\citep{hendrycks2021measuring},
TruthfulQA~\citep{lin2022truthfulqa}, and
Winogrande~\citep{sakaguchi2019winogrande}. We selected these evaluations based on their use as English language benchmarks by \newcite{open-llm-leaderboard} and use an identical testing configuration here.
Programming language proficiency is assessed via the Bigcode Evaluation Harness~\citep{bigcode-evaluation-harness} with the
HumanEval~\citep{chen2021evaluating}, and
MBPP~\citep{austin2021program} benchmarks, employing the pass@10 metric for evaluation.

To evaluate the quality of Finnish text generation, we generate responses to the translated MT-Bench questions with few-shot prompting~\cite{zheng2023judging}. We use a few-shot prompt because this benchmark is designed for chat models and we are evaluating base models. Moreover, we want to unlock the Finnish generation capabilities of the English-focused models by providing in-context examples in Finnish. We use GPT-4 Turbo and human judges to assess the quality of the responses. Finally, to evaluate translation performance, we use both the Flores-101 devtest~\citep{goyal2022flores} as well as the Tatoeba test sets~\citep{tiedemann2020tatoeba} in an 8-shot setting, following \newcite{zhu2023multilingual}.

\subsection{Perplexity}
\label{sec:eval-results-cppl}

\begin{table*}[t!]
\centering
\setlength{\tabcolsep}{3pt}
\begin{tabular}{l|c|c c c| c c |c}
  & \bf Poro~34B & \bf Llama 33B & \bf MPT 30b & \bf Falcon 40B & \bf FinGPT 8B & \bf FinGPT 13B & \bf Starcoder \\
\hline
Finnish  & \bf 66.28 & 53.36 & 53.22 & 42.58 & 49.69 & 48.92 & 45.55 \\
English  & 50.57 & \bf 59.96 & 52.62 & 49.87 & 31.47 & 32.85 & 35.44 \\
Code & 41.80 & 37.67 & 39.18 & 38.57 & - & - & \bf 49.06\\
\end{tabular}
\caption{Average benchmark results for Finnish, English and code for Poro~34B and selected reference models.}
\label{tab:eval-results}
\end{table*}

Table~\ref{tab:char-ppl} summarizes the results of the perplexity evaluation as mean character-level perplexity $PPL_c$ for various models over the sentences/code lines. We find that Poro~34B has comparatively low (good) $PPL_c$ on all three datasets, including the best result for Finnish. Poro~34B is to the best of our knowledge the only open model specifically trained for this combination of languages, and it is thus not surprising that it has the best overall average in this evaluation. While perplexity is not necessarily predictive of downstream performance and these datasets only represent a part of the relevant distribution, the result suggests that the model has learned all of its target languages well.

\begin{figure}[ht!]
\centering
\includegraphics[width=\columnwidth]{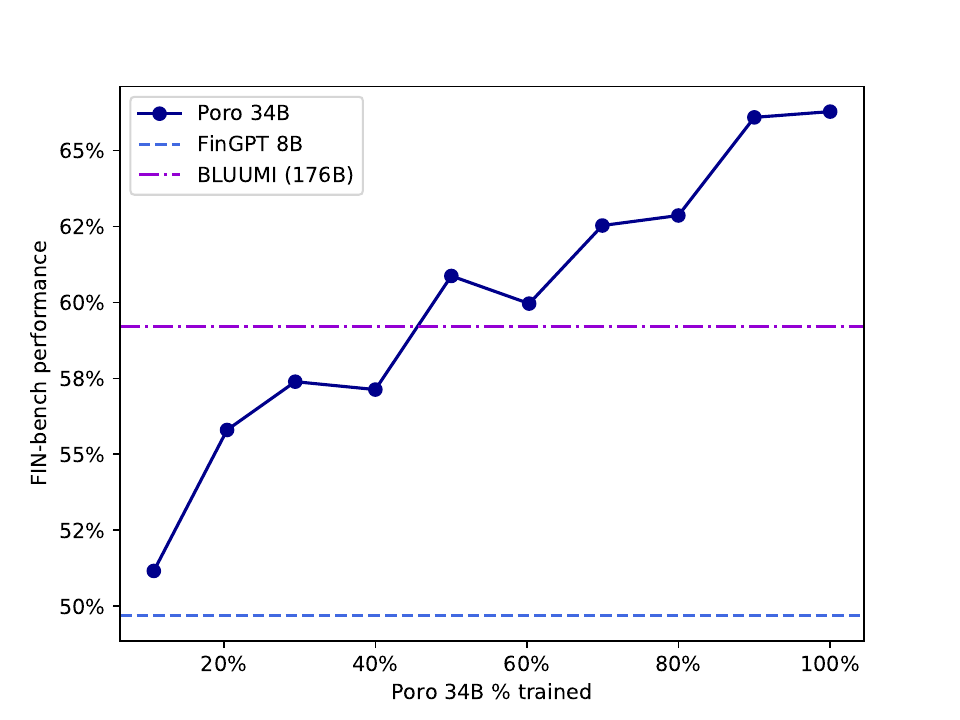}
\caption{Poro~34B performance progression on FIN-bench. For reference, dotted lines show results for the best-performing monolingual FinGPT model and the massively multilingual BLUUMI model \citep{luukkonen2023fingpt}, an extension of BLOOM \citep{lescao2022bloom} with Finnish.}
\label{fig:poro-fingpt-comparison}
\end{figure}

\subsection{Benchmark results}
\label{sec:eval-results-tasks}

The overall results of the benchmark evaluations are summarized in Table~\ref{tab:eval-results} and detailed in Appendix~\ref{sec:benchmark_details}.
We find that Poro~34B is the best-performing model for Finnish in this comparison, substantially outperforming the best previously introduced monolingual Finnish model. We further analyzed the progression of the Finnish capabilities by evaluating Poro~34B checkpoints at 10\% intervals on FIN-bench. These results are summarized in Figure~\ref{fig:poro-fingpt-comparison}. 
Interestingly, the model outperforms the best FinGPT model already after 100B tokens of training (10\%) despite the relatively small proportion of Finnish in the Poro~34B data and the fact that the FinGPT models were trained on 300B tokens in total.
These results indicate that our limited multilingual approach is effective for creating stronger models for Finnish than possible through monolingual training and demonstrate that the model is benefiting substantially from its training data in other languages even when tested on Finnish tasks.

For English, we find that the model achieves broadly comparable results to the MPT 30B and Falcon 40B models, both of which were trained for 1T tokens of predominantly English data. This indicates that the limited multilingual training approach has not notably detracted from the English capabilities of the model. The best-performing open model in this comparison is Llama~33B, which was trained for longer (1.4T tokens), also predominantly on English data. We find that Poro~34B is nevertheless a capable model in its class also for English, despite not optimizing specifically for English performance. The programming language benchmarks indicate that Poro~34B is more capable on code than the other natural language-focused models, while the code-focused StarCoder model clearly outperforms all of the other models. We attribute the relatively high performance of Poro~34B on code to the comparatively large proportion of the training data dedicated to code. As with English, we consider the performance of the model on code a positive addition even though code generation was not a primary goal in creating the model.

\begin{figure*}
\centering
\includegraphics[width=0.9\textwidth]{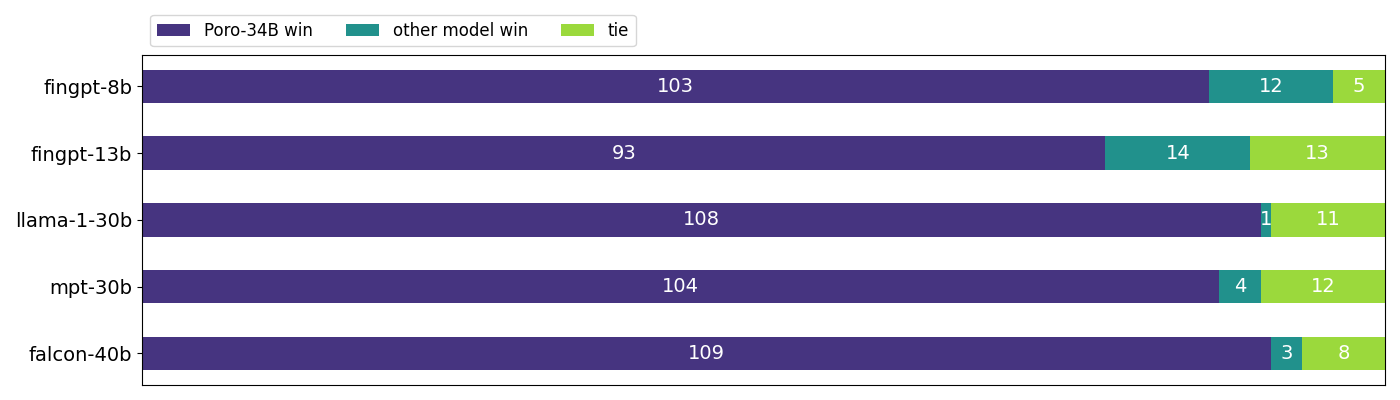}
\caption{Win counts of reference models against Poro~34B on Finnish MT-Bench as judged by GPT-4 Turbo.}
\label{fig:mtbench_results_gpt}
\end{figure*}

Finally, we note a surprising finding arising from the Finnish evaluation: two of the larger English-focused models (Llama 33B and MPT 30B) score higher than the previously introduced smaller monolingual Finnish models on the FIN-bench benchmark. While FIN-bench tasks are in Finnish, the benchmark consists of multiple-choice rather than generation tasks, has been produced in substantial part through translation from English, and includes tasks with little emphasis on natural language (esp.\ arithmetic). We hypothesize that the comparatively high performance of the English-focused models on this benchmark might not indicate that they can generate fluent Finnish, which also calls the Finnish proficiency of Poro~34B into question. We study this question specifically in the following section.

\subsection{Open-ended generation}
\label{sec:finnish-generation}

To assess the ability of the models to generate coherent and grammatically correct Finnish, we create a Finnish version of MT-Bench~\cite{zheng2023judging}, a benchmark for open-ended conversations that uses LLM-as-a-judge evaluation. We excluded math and coding questions to focus specifically on the natural language generation capabilities of the models. To create the benchmark, we initially translated the questions into Finnish using DeepL,\footnote{\url{https://www.deepl.com}} and the translations were then manually corrected by native Finnish speakers to create the final evaluation dataset. To evaluate base models using the data, we similarly translated and corrected the few-shot URIAL prompt~\cite{Lin2024ReAlign}.
\footnote{
We did not modify the judge prompts as previous work has found that keeping the prompt in English produces better results~\cite{ahuja2023mega}.}
We use pairwise judging to compare between Poro~34B and the competing models' responses and use GPT-4 Turbo as the judge model.

To assess the reliability of the model as a judge and provide further insight into the quality of the generations, we additionally set up an annotation platform where two native Finnish speakers were asked to pick a preference between a response generated by Poro~34B and a competing model.\footnote{We did not separately compensate the human judges as they are co-authors of this paper.} The judges are given the same judging prompt as GPT.  The model names are hidden from the judges, and we randomly select the position of each response in every response pair  to account for positional bias.

We found that the two human judges highly agree with each other, picking the same winner 88.8\% of the time, and found an even higher agreement between GPT and each human judge: 91.6\% between annotator 1 and GPT and 89.5\% between annotator 2 and GPT. Figure~\ref{fig:mtbench_results_gpt} shows the win counts of the reference models against Poro~34B as judged by GPT-4 Turbo.

In manual analysis after the initial annotation, we found that the FinGPT models often struggled with the few-shot format, failing to follow questions or only giving short, minimal answers, while Poro~34B was better able to comply with questions and given requirements, such as listing a specified number of items. However, we found that Poro~34B also often hallucinated and did not follow all instructions, and we would not consider its responses to be at a level of consistency and quality required for user-facing applications, which is not an unexpected result given that it is a base model not specifically fine-tuned or otherwise aligned for such use. 
Despite outperforming FinGPT models on the FIN-bench benchmark, The English-focused models appeared to be unfit for Finnish generation: their generations had the surface appearance of Finnish text but were largely nonsensical and incoherent. This result underlines the need to include multiple perspectives when evaluating models: a high score on a multiple-choice benchmark may not indicate practical capability to generate coherent text in a language.

We make the Finnish MT-Bench available under an open license and provide the model generations at \url{https://github.com/LumiOpen/FastChat/tree/main/fastchat/llm_judge}.

\subsection{Translation}

\interfootnotelinepenalty=99999

General-purpose language models have shown promising results on translation benchmarks on multiple languages~\citep{vilar2023prompting,garcia2023unreasonable,alves2024tower}. Following \citet{zhu2023multilingual}, we evaluated Poro~34B for English to Finnish translation and vice versa on the first 100 sentences of the Flores-101 test data by prompting the model with eight translation examples sampled randomly from the development set, formatting the examples simply as \texttt{<src>=<trg>}.
We further evaluated Poro~34B and three strong open-source translation models on the Tatoeba test set with more than 11,000 sentences: OPUS-MT~\citep{TiedemannThottingal:EAMT2020}, NLLB-1.3B~\citep{costa2022no}, and M2M-100-12B~\citep{fan2021beyond}\footnote{We did not evaluate the GPT models and Google Translate on Tatoeba because of the associated API costs.}. We used the standard SentencePiece BLEU (spBLEU) as our metric.
The results of both evaluations are shown in Table~\ref{tab:translation-evals}.\footnote{We attempted to reproduce some of the Flores-101 results reported by \cite{zhu2023multilingual} and obtained a slightly higher result for GPT-4 in Eng-Fin translation (37.5 instead of 35.33) and slightly lower results for M2M-12B and NLLB-1.3B (31.4 and 26.6, respectively). For the sake of consistency, we present the results from that study without modification.} These results demonstrate that Poro~34B is a remarkably strong translator, outperforming not only dedicated open-source translation models but even Google Translate, and scoring roughly on par with GPT-4 in this evaluation. We attribute this result to the combination of strong Finnish and English capabilities and the inclusion of a comparatively large number of translation examples in the pretraining data.

\begin{table}[t!]
\centering
\begin{tabular}{ l c c | c c}
 & \multicolumn{2}{c}{\textbf{Flores-101}} & \multicolumn{2}{c}{\textbf{Tatoeba}}\\
\textbf{Model} & \textbf{En-Fi} & \textbf{Fi-En} & \textbf{En-Fi} & \textbf{Fi-En} \\
\hline
ChatGPT	& 33.4 & 35.9 & {-} & {-} \\
GPT-4	& 35.3 & \textbf{40.2} & {-} & {-} \\
Google	& 37.3 & 39.0 & {-} & {-} \\ 
M2M-12B	& 33.3 & 33.8 & 36.7 & 41.3 \\
NLLB-1.3B	& 30.0 & 35.4 & 40.2 & 55.7 \\
OPUS-MT	& {37.2} & 35.6 & {46.7} & {58.4} \\
Poro~34B & \textbf{37.6} & {39.8} & \textbf{47.3} & \textbf{60.5}\\ 
\end{tabular}
\caption{spBLEU on the Flores-101 devtest and Tatoeba test sets. Flores-101 results except for OPUS-MT and Poro~34B are from~\citet{zhu2023multilingual}.}
\label{tab:translation-evals}
\end{table}

It should be noted, however, that the Tatoeba and Flores sentences are relatively short and simple, and this evaluation does thus not capture the full picture of the translation capabilities of the evaluated models. We aim to assess the translation capability of Poro~34B more comprehensively on longer texts, especially texts that might include different modalities such as tables and code, in future work.

\section{Discussion and conclusions}

In this study, we have considered the challenges that the availability of data poses for pretraining large generative models for smaller languages and explored a limited multilingual approach to create Poro~34B, a 34B-parameter model trained on 1T tokens of Finnish, English, and code, including 8B tokens of Finnish-English translation pairs. We thoroughly evaluated the model and found it to substantially advance over the performance of existing models for Finnish while also performing competitively in its class of open models for English and code generation, as well as achieving remarkably good results in translation tasks. Two human judges and GPT-4 Turbo found the texts generated by Poro~34B to be superior to the competing models. 


Our model architecture and the Finnish datasets included follow those of the FinGPT family of monolingual Finnish models, which were constrained by the available Finnish training data. The superior performance of our model in Finnish evaluations demonstrates that multilingual training can lift such limitations, allowing further scaling of models focused on smaller languages. In future work, we hope to explore this effect more systematically to answer some of the many questions that remain open regarding the training of large generative models for smaller languages, including the impacts of covering multiple smaller languages and the effect of the size of data available in the target languages.

A number of the choices made in training Poro~34B were made with incomplete information regarding their specific impacts on the final model. For example, we opted to include a comparatively large amount of programming language data as well as instruction-formatted translation examples in the pretraining data, the latter on the assumption that this would provide a cross-lingual signal that would strengthen the ability of the model to benefit from data in a more distantly related language (English). While this approach is intuitively appealing and the performance of our model suggests that it has at a minimum not notably detracted from the capabilities of the model, we did not as part of this work have the resources to conduct ablation studies nor to explore alternative ways to incorporate cross-lingual information in pretraining. We aim to study these questions further in future work.

We hope that our approach can serve as a template for the creation of larger models for other smaller languages and that the model introduced in this work can serve as both as a focus of research in its own right as well as a starting point for further pretraining, finetuning and alignment to create useful models, tools and methods not only for Finnish but also other languages. We release the model weights as well as all relevant documentation and software fully openly at \url{https://huggingface.co/LumiOpen/Poro-34B}.

\subsection*{Limitations}

Our study applies a pretraining recipe that combines insights on effective multilingual and data-constrained model training from a variety of previous studies. While the findings of these studies are supported by a broad range of relevant experimental results, we did not have the resources to perform separate ablation experiments specifically assessing the impact that various parts of our combined pretraining recipe (e.g., four repetitions of target language data and the inclusion of a translation signal) have on the resulting model. Thus, while we believe that our results demonstrate the pretraining recipe to be effective for creating state-of-the-art models for data-constrained languages, our work is limited in leaving many questions open regarding specific choices that form part of that recipe.


Poro~34B is a base model and as such has not been aligned to follow instructions and engage in conversations. It has not been evaluated on safety and toxicity benchmarks. As we have noted in our language generation evaluation, Poro~34B does not adequately follow instructions and has the tendency to generate texts with hallucinations. Further research is needed to improve the model in terms of factuality, safety, and alignment in English and Finnish.
We encourage developers using Poro~34B to be aware of the potential risks associated with LLMs such as non-factual outputs, harmful language, and perpetuation of biases and stereotypes. We recommend that developers finetune Poro~34B to meet their specific needs and codes of conduct.

\subsection*{Ethical considerations}

We are committed to open science, transparency and accessibility in our work. While we acknowledge the concerns and the potential for negative impacts associated with making powerful generative models and the technology to create them more widely available, we believe that in the case of Poro~34B the positives clearly outweigh the negatives. We discuss some specific concerns and their mitigations in the following.

Poro~34B is a base model trained in substantial part on texts sourced from web crawls, which are known to include biases, toxicity and factual errors. While we have selected curated text sources that have been extensively filtered to remove problematic material, no such filtering is perfect. Like all language models, Poro~34B is a product of its inputs, and its output may reflect issues in its training material. Furthermore, as Poro~34B is a base model that has not been finetuned for any specific purpose, extra care should be taken when interpreting its output, and the model should not be used as is in any application with potential for significant impact on people’s rights or well-being. We emphasize these limitations in the model card published with the model.

Pretraining large language models is computationally intensive, and the creation of large models can have substantial environmental impacts. Poro~34B was trained on the LUMI supercomputer, which is powered entirely by renewable energy resources. According to the official specifications, the carbon intensity factor of LUMI's operation is considered to be zero. This approach effectively minimizes the carbon footprint associated with the computational aspects of training our model.

Though concerns about the capabilities of frontier models to cause catastrophic harm have been discussed in the literature, a model of Poro~34B's size and training duration does not represent new frontier capability and releasing the model does not introduce any new classes of risk.

\subsection*{Acknowledgements}
The authors wish to acknowledge CSC – IT Center for Science, Finland, for generous computational resources on the LUMI supercomputer. This project has received funding from the European Union’s Horizon Europe research and innovation programme under Grant agreement No 101070350. The contents of this publication are the sole responsibility of its authors and do not necessarily reflect the opinion of the European Union.

\bibliographystyle{acl_nodalida}
\bibliography{acl_nodalida}

\begin{thebibliography}{69}
\expandafter\ifx\csname natexlab\endcsname\relax\def\natexlab#1{#1}\fi

\bibitem[{Ahuja et~al.(2023)Ahuja, Diddee, Hada, Ochieng, Ramesh, Jain, Nambi, Ganu, Segal, Ahmed et~al.}]{ahuja2023mega}
Kabir Ahuja, Harshita Diddee, Rishav Hada, Millicent Ochieng, Krithika Ramesh, Prachi Jain, Akshay Nambi, Tanuja Ganu, Sameer Segal, Mohamed Ahmed, et~al. 2023.
\newblock Mega: Multilingual evaluation of generative ai.
\newblock In \emph{Proceedings of the 2023 Conference on Empirical Methods in Natural Language Processing}, pages 4232--4267.

\bibitem[{Al-Khateeb et~al.(2023)Al-Khateeb, Dey, Soboleva, and Hestness}]{al2023position}
Faisal Al-Khateeb, Nolan Dey, Daria Soboleva, and Joel Hestness. 2023.
\newblock {Position Interpolation Improves ALiBi Extrapolation}.
\newblock \emph{arXiv preprint arXiv:2310.13017}.

\bibitem[{Ali et~al.(2023)Ali, Fromm, Thellmann, Rutmann, L{\"u}bbering, Leveling, Klug, Ebert, Doll, Buschhoff et~al.}]{ali2023tokenizer}
Mehdi Ali, Michael Fromm, Klaudia Thellmann, Richard Rutmann, Max L{\"u}bbering, Johannes Leveling, Katrin Klug, Jan Ebert, Niclas Doll, Jasper~Schulze Buschhoff, et~al. 2023.
\newblock {Tokenizer Choice For LLM Training: Negligible or Crucial?}
\newblock \emph{arXiv preprint arXiv:2310.08754}.

\bibitem[{Almazrouei et~al.(2023)Almazrouei, Alobeidli, Alshamsi, Cappelli, Cojocaru, Debbah, Goffinet, Hesslow, Launay, Malartic et~al.}]{almazrouei2023falcon}
Ebtesam Almazrouei, Hamza Alobeidli, Abdulaziz Alshamsi, Alessandro Cappelli, Ruxandra Cojocaru, M{\'e}rouane Debbah, {\'E}tienne Goffinet, Daniel Hesslow, Julien Launay, Quentin Malartic, et~al. 2023.
\newblock {The {Falcon} series of open language models}.
\newblock \emph{arXiv preprint arXiv:2311.16867}.

\bibitem[{Alves et~al.(2024)Alves, Pombal, Guerreiro, Martins, Alves, Farajian, Peters, Rei, Fernandes, Agrawal, Colombo, de~Souza, and Martins}]{alves2024tower}
Duarte~M. Alves, José Pombal, Nuno~M. Guerreiro, Pedro~H. Martins, João Alves, Amin Farajian, Ben Peters, Ricardo Rei, Patrick Fernandes, Sweta Agrawal, Pierre Colombo, José G.~C. de~Souza, and André F.~T. Martins. 2024.
\newblock \href {http://arxiv.org/abs/2402.17733} {Tower: An open multilingual large language model for translation-related tasks}.

\bibitem[{Anil et~al.(2023)Anil, Dai, Firat, Johnson, Lepikhin, Passos, Shakeri, Taropa, Bailey, Chen et~al.}]{anil2023palm}
Rohan Anil, Andrew~M Dai, Orhan Firat, Melvin Johnson, Dmitry Lepikhin, Alexandre Passos, Siamak Shakeri, Emanuel Taropa, Paige Bailey, Zhifeng Chen, et~al. 2023.
\newblock Palm 2 technical report.
\newblock \emph{arXiv preprint arXiv:2305.10403}.

\bibitem[{Aryabumi et~al.(2024)Aryabumi, Su, Ma, Morisot, Zhang, Locatelli, Fadaee, Üstün, and Hooker}]{aryabumi2024code}
Viraat Aryabumi, Yixuan Su, Raymond Ma, Adrien Morisot, Ivan Zhang, Acyr Locatelli, Marzieh Fadaee, Ahmet Üstün, and Sara Hooker. 2024.
\newblock \href {http://arxiv.org/abs/2408.10914} {To code, or not to code? exploring impact of code in pre-training}.

\bibitem[{Austin et~al.(2021)Austin, Odena, Nye, Bosma, Michalewski, Dohan, Jiang, Cai, Terry, Le et~al.}]{austin2021program}
Jacob Austin, Augustus Odena, Maxwell Nye, Maarten Bosma, Henryk Michalewski, David Dohan, Ellen Jiang, Carrie Cai, Michael Terry, Quoc Le, et~al. 2021.
\newblock Program synthesis with large language models.
\newblock \emph{arXiv preprint arXiv:2108.07732}.

\bibitem[{Beeching et~al.(2023)Beeching, Fourrier, Habib, Han, Lambert, Rajani, Sanseviero, Tunstall, and Wolf}]{open-llm-leaderboard}
Edward Beeching, Clémentine Fourrier, Nathan Habib, Sheon Han, Nathan Lambert, Nazneen Rajani, Omar Sanseviero, Lewis Tunstall, and Thomas Wolf. 2023.
\newblock Open llm leaderboard.
\newblock \url{https://huggingface.co/spaces/HuggingFaceH4/open_llm_leaderboard}.

\bibitem[{Ben~Allal et~al.(2022)Ben~Allal, Muennighoff, Kumar~Umapathi, Lipkin, and von Werra}]{bigcode-evaluation-harness}
Loubna Ben~Allal, Niklas Muennighoff, Logesh Kumar~Umapathi, Ben Lipkin, and Leandro von Werra. 2022.
\newblock A framework for the evaluation of code generation models.
\newblock \url{https://github.com/bigcode-project/bigcode-evaluation-harness}.

\bibitem[{Brown et~al.(2020)Brown, Mann, Ryder, Subbiah, Kaplan, Dhariwal, Neelakantan, Shyam, Sastry, Askell, Agarwal, Herbert-Voss, Krueger, Henighan, Child, Ramesh, Ziegler, Wu, Winter, Hesse, Chen, Sigler, Litwin, Gray, Chess, Clark, Berner, McCandlish, Radford, Sutskever, and Amodei}]{brown2020language}
Tom Brown, Benjamin Mann, Nick Ryder, Melanie Subbiah, Jared~D Kaplan, Prafulla Dhariwal, Arvind Neelakantan, Pranav Shyam, Girish Sastry, Amanda Askell, Sandhini Agarwal, Ariel Herbert-Voss, Gretchen Krueger, Tom Henighan, Rewon Child, Aditya Ramesh, Daniel Ziegler, Jeffrey Wu, Clemens Winter, Chris Hesse, Mark Chen, Eric Sigler, Mateusz Litwin, Scott Gray, Benjamin Chess, Jack Clark, Christopher Berner, Sam McCandlish, Alec Radford, Ilya Sutskever, and Dario Amodei. 2020.
\newblock \href {https://proceedings.neurips.cc/paper_files/paper/2020/file/1457c0d6bfcb4967418bfb8ac142f64a-Paper.pdf} {Language models are few-shot learners}.
\newblock In \emph{Advances in Neural Information Processing Systems}, volume~33, pages 1877--1901. Curran Associates, Inc.

\bibitem[{Chang et~al.(2023)Chang, Arnett, Tu, and Bergen}]{chang2023multilinguality}
Tyler~A. Chang, Catherine Arnett, Zhuowen Tu, and Benjamin~K. Bergen. 2023.
\newblock \href {http://arxiv.org/abs/2311.09205} {When is multilinguality a curse? language modeling for 250 high- and low-resource languages}.

\bibitem[{Chen et~al.(2021)Chen, Tworek, Jun, Yuan, de~Oliveira~Pinto, Kaplan, Edwards, Burda, Joseph, Brockman, Ray, Puri, Krueger, Petrov, Khlaaf, Sastry, Mishkin, Chan, Gray, Ryder, Pavlov, Power, Kaiser, Bavarian, Winter, Tillet, Such, Cummings, Plappert, Chantzis, Barnes, Herbert-Voss, Guss, Nichol, Paino, Tezak, Tang, Babuschkin, Balaji, Jain, Saunders, Hesse, Carr, Leike, Achiam, Misra, Morikawa, Radford, Knight, Brundage, Murati, Mayer, Welinder, McGrew, Amodei, McCandlish, Sutskever, and Zaremba}]{chen2021evaluating}
Mark Chen, Jerry Tworek, Heewoo Jun, Qiming Yuan, Henrique~Ponde de~Oliveira~Pinto, Jared Kaplan, Harri Edwards, Yuri Burda, Nicholas Joseph, Greg Brockman, Alex Ray, Raul Puri, Gretchen Krueger, Michael Petrov, Heidy Khlaaf, Girish Sastry, Pamela Mishkin, Brooke Chan, Scott Gray, Nick Ryder, Mikhail Pavlov, Alethea Power, Lukasz Kaiser, Mohammad Bavarian, Clemens Winter, Philippe Tillet, Felipe~Petroski Such, Dave Cummings, Matthias Plappert, Fotios Chantzis, Elizabeth Barnes, Ariel Herbert-Voss, William~Hebgen Guss, Alex Nichol, Alex Paino, Nikolas Tezak, Jie Tang, Igor Babuschkin, Suchir Balaji, Shantanu Jain, William Saunders, Christopher Hesse, Andrew~N. Carr, Jan Leike, Josh Achiam, Vedant Misra, Evan Morikawa, Alec Radford, Matthew Knight, Miles Brundage, Mira Murati, Katie Mayer, Peter Welinder, Bob McGrew, Dario Amodei, Sam McCandlish, Ilya Sutskever, and Wojciech Zaremba. 2021.
\newblock \href {http://arxiv.org/abs/2107.03374} {Evaluating large language models trained on code}.

\bibitem[{Clark et~al.(2018)Clark, Cowhey, Etzioni, Khot, Sabharwal, Schoenick, and Tafjord}]{Clark2018ThinkYH}
Peter Clark, Isaac Cowhey, Oren Etzioni, Tushar Khot, Ashish Sabharwal, Carissa Schoenick, and Oyvind Tafjord. 2018.
\newblock {Think you have Solved Question Answering? Try ARC, the AI2 Reasoning Challenge}.
\newblock \emph{ArXiv}, abs/1803.05457.

\bibitem[{Cobbe et~al.(2021)Cobbe, Kosaraju, Bavarian, Chen, Jun, Kaiser, Plappert, Tworek, Hilton, Nakano, Hesse, and Schulman}]{cobbe2021training}
Karl Cobbe, Vineet Kosaraju, Mohammad Bavarian, Mark Chen, Heewoo Jun, Lukasz Kaiser, Matthias Plappert, Jerry Tworek, Jacob Hilton, Reiichiro Nakano, Christopher Hesse, and John Schulman. 2021.
\newblock \href {http://arxiv.org/abs/2110.14168} {Training verifiers to solve math word problems}.

\bibitem[{Conneau et~al.(2020)Conneau, Khandelwal, Goyal, Chaudhary, Wenzek, Guzm{\'a}n, Grave, Ott, Zettlemoyer, and Stoyanov}]{conneau2020unsupervised}
Alexis Conneau, Kartikay Khandelwal, Naman Goyal, Vishrav Chaudhary, Guillaume Wenzek, Francisco Guzm{\'a}n, Edouard Grave, Myle Ott, Luke Zettlemoyer, and Veselin Stoyanov. 2020.
\newblock \href {https://doi.org/10.18653/v1/2020.acl-main.747} {Unsupervised cross-lingual representation learning at scale}.
\newblock In \emph{Proceedings of the 58th Annual Meeting of the Association for Computational Linguistics}, pages 8440--8451. Association for Computational Linguistics.

\bibitem[{Costa-juss{\`a} et~al.(2022)Costa-juss{\`a}, Cross, {\c{C}}elebi, Elbayad, Heafield, Heffernan, Kalbassi, Lam, Licht, Maillard et~al.}]{costa2022no}
Marta~R Costa-juss{\`a}, James Cross, Onur {\c{C}}elebi, Maha Elbayad, Kenneth Heafield, Kevin Heffernan, Elahe Kalbassi, Janice Lam, Daniel Licht, Jean Maillard, et~al. 2022.
\newblock No language left behind: Scaling human-centered machine translation.
\newblock \emph{arXiv preprint arXiv:2207.04672}.

\bibitem[{Devlin et~al.(2019)Devlin, Chang, Lee, and Toutanova}]{devlin2019bert}
Jacob Devlin, Ming-Wei Chang, Kenton Lee, and Kristina Toutanova. 2019.
\newblock \href {https://doi.org/10.18653/v1/N19-1423} {{BERT}: Pre-training of deep bidirectional transformers for language understanding}.
\newblock In \emph{Proceedings of the 2019 Conference of the North {A}merican Chapter of the Association for Computational Linguistics: Human Language Technologies, Volume 1 (Long and Short Papers)}, pages 4171--4186, Minneapolis, Minnesota. Association for Computational Linguistics.

\bibitem[{Ekgren et~al.(2022)Ekgren, Cuba~Gyllensten, Gogoulou, Heiman, Verlinden, {\"O}hman, Carlsson, and Sahlgren}]{ekgren2022lessons}
Ariel Ekgren, Amaru Cuba~Gyllensten, Evangelia Gogoulou, Alice Heiman, Severine Verlinden, Joey {\"O}hman, Fredrik Carlsson, and Magnus Sahlgren. 2022.
\newblock \href {https://aclanthology.org/2022.lrec-1.376} {Lessons learned from {GPT}-{SW}3: Building the first large-scale generative language model for {S}wedish}.
\newblock In \emph{Proceedings of the Thirteenth Language Resources and Evaluation Conference}, pages 3509--3518.

\bibitem[{Fan et~al.(2021)Fan, Bhosale, Schwenk, Ma, El-Kishky, Goyal, Baines, Celebi, Wenzek, Chaudhary et~al.}]{fan2021beyond}
Angela Fan, Shruti Bhosale, Holger Schwenk, Zhiyi Ma, Ahmed El-Kishky, Siddharth Goyal, Mandeep Baines, Onur Celebi, Guillaume Wenzek, Vishrav Chaudhary, et~al. 2021.
\newblock Beyond english-centric multilingual machine translation.
\newblock \emph{Journal of Machine Learning Research}, 22(107):1--48.

\bibitem[{Fujinuma et~al.(2022)Fujinuma, Boyd-Graber, and Kann}]{fujinuma2022match}
Yoshinari Fujinuma, Jordan Boyd-Graber, and Katharina Kann. 2022.
\newblock \href {https://doi.org/10.18653/v1/2022.acl-long.106} {Match the script, adapt if multilingual: Analyzing the effect of multilingual pretraining on cross-lingual transferability}.
\newblock In \emph{Proceedings of the 60th Annual Meeting of the Association for Computational Linguistics (Volume 1: Long Papers)}, pages 1500--1512. Association for Computational Linguistics.

\bibitem[{Gao et~al.(2023)Gao, Tow, Abbasi, Biderman, Black, DiPofi, Foster, Golding, Hsu, Le~Noac'h, Li, McDonell, Muennighoff, Ociepa, Phang, Reynolds, Schoelkopf, Skowron, Sutawika, Tang, Thite, Wang, Wang, and Zou}]{eval-harness}
Leo Gao, Jonathan Tow, Baber Abbasi, Stella Biderman, Sid Black, Anthony DiPofi, Charles Foster, Laurence Golding, Jeffrey Hsu, Alain Le~Noac'h, Haonan Li, Kyle McDonell, Niklas Muennighoff, Chris Ociepa, Jason Phang, Laria Reynolds, Hailey Schoelkopf, Aviya Skowron, Lintang Sutawika, Eric Tang, Anish Thite, Ben Wang, Kevin Wang, and Andy Zou. 2023.
\newblock \href {https://doi.org/10.5281/zenodo.10256836} {A framework for few-shot language model evaluation}.

\bibitem[{Garcia et~al.(2023)Garcia, Bansal, Cherry, Foster, Krikun, Johnson, and Firat}]{garcia2023unreasonable}
Xavier Garcia, Yamini Bansal, Colin Cherry, George Foster, Maxim Krikun, Melvin Johnson, and Orhan Firat. 2023.
\newblock The unreasonable effectiveness of few-shot learning for machine translation.
\newblock In \emph{International Conference on Machine Learning}, pages 10867--10878. PMLR.

\bibitem[{Gogoulou et~al.(2023)Gogoulou, Lesort, Boman, and Nivre}]{gogoulou2023continual}
Evangelia Gogoulou, Timothée Lesort, Magnus Boman, and Joakim Nivre. 2023.
\newblock \href {http://arxiv.org/abs/2311.01200} {Continual learning under language shift}.

\bibitem[{Goyal et~al.(2022)Goyal, Gao, Chaudhary, Chen, Wenzek, Ju, Krishnan, Ranzato, Guzm{\'a}n, and Fan}]{goyal2022flores}
Naman Goyal, Cynthia Gao, Vishrav Chaudhary, Peng-Jen Chen, Guillaume Wenzek, Da~Ju, Sanjana Krishnan, Marc’Aurelio Ranzato, Francisco Guzm{\'a}n, and Angela Fan. 2022.
\newblock The flores-101 evaluation benchmark for low-resource and multilingual machine translation.
\newblock \emph{Transactions of the Association for Computational Linguistics}, 10:522--538.

\bibitem[{Groeneveld et~al.(2024)Groeneveld, Beltagy, Walsh, Bhagia, Kinney, Tafjord, Jha, Ivison, Magnusson, Wang, Arora, Atkinson, Authur, Chandu, Cohan, Dumas, Elazar, Gu, Hessel, Khot, Merrill, Morrison, Muennighoff, Naik, Nam, Peters, Pyatkin, Ravichander, Schwenk, Shah, Smith, Strubell, Subramani, Wortsman, Dasigi, Lambert, Richardson, Zettlemoyer, Dodge, Lo, Soldaini, Smith, and Hajishirzi}]{groeneveld2024olmo}
Dirk Groeneveld, Iz~Beltagy, Pete Walsh, Akshita Bhagia, Rodney Kinney, Oyvind Tafjord, Ananya~Harsh Jha, Hamish Ivison, Ian Magnusson, Yizhong Wang, Shane Arora, David Atkinson, Russell Authur, Khyathi~Raghavi Chandu, Arman Cohan, Jennifer Dumas, Yanai Elazar, Yuling Gu, Jack Hessel, Tushar Khot, William Merrill, Jacob Morrison, Niklas Muennighoff, Aakanksha Naik, Crystal Nam, Matthew~E. Peters, Valentina Pyatkin, Abhilasha Ravichander, Dustin Schwenk, Saurabh Shah, Will Smith, Emma Strubell, Nishant Subramani, Mitchell Wortsman, Pradeep Dasigi, Nathan Lambert, Kyle Richardson, Luke Zettlemoyer, Jesse Dodge, Kyle Lo, Luca Soldaini, Noah~A. Smith, and Hannaneh Hajishirzi. 2024.
\newblock \href {http://arxiv.org/abs/2402.00838} {{OLMo}: Accelerating the science of language models}.

\bibitem[{Hendrycks et~al.(2021)Hendrycks, Burns, Basart, Zou, Mazeika, Song, and Steinhardt}]{hendrycks2021measuring}
Dan Hendrycks, Collin Burns, Steven Basart, Andy Zou, Mantas Mazeika, Dawn Song, and Jacob Steinhardt. 2021.
\newblock Measuring massive multitask language understanding.
\newblock \emph{Proceedings of the International Conference on Learning Representations (ICLR)}.

\bibitem[{Hernandez et~al.(2022)Hernandez, Brown, Conerly, DasSarma, Drain, El-Showk, Elhage, Hatfield-Dodds, Henighan, Hume et~al.}]{hernandez2022scaling}
Danny Hernandez, Tom Brown, Tom Conerly, Nova DasSarma, Dawn Drain, Sheer El-Showk, Nelson Elhage, Zac Hatfield-Dodds, Tom Henighan, Tristan Hume, et~al. 2022.
\newblock Scaling laws and interpretability of learning from repeated data.
\newblock \emph{arXiv preprint arXiv:2205.10487}.

\bibitem[{Hoffmann et~al.(2022{\natexlab{a}})Hoffmann, Borgeaud, Mensch, Buchatskaya, Cai, Rutherford, Casas, Hendricks, Welbl, Clark et~al.}]{hoffmann2022training}
Jordan Hoffmann, Sebastian Borgeaud, Arthur Mensch, Elena Buchatskaya, Trevor Cai, Eliza Rutherford, Diego de~Las Casas, Lisa~Anne Hendricks, Johannes Welbl, Aidan Clark, et~al. 2022{\natexlab{a}}.
\newblock Training compute-optimal large language models.
\newblock \emph{arXiv preprint arXiv:2203.15556}.

\bibitem[{Hoffmann et~al.(2022{\natexlab{b}})Hoffmann, Borgeaud, Mensch, Buchatskaya, Cai, Rutherford, de~Las~Casas, Hendricks, Welbl, Clark, Hennigan, Noland, Millican, van~den Driessche, Damoc, Guy, Osindero, Simonyan, Elsen, Vinyals, Rae, and Sifre}]{hoffmann2022empirical}
Jordan Hoffmann, Sebastian Borgeaud, Arthur Mensch, Elena Buchatskaya, Trevor Cai, Eliza Rutherford, Diego de~Las~Casas, Lisa~Anne Hendricks, Johannes Welbl, Aidan Clark, Thomas Hennigan, Eric Noland, Katherine Millican, George van~den Driessche, Bogdan Damoc, Aurelia Guy, Simon Osindero, Kar\'{e}n Simonyan, Erich Elsen, Oriol Vinyals, Jack Rae, and Laurent Sifre. 2022{\natexlab{b}}.
\newblock \href {https://proceedings.neurips.cc/paper_files/paper/2022/file/c1e2faff6f588870935f114ebe04a3e5-Paper-Conference.pdf} {An empirical analysis of compute-optimal large language model training}.
\newblock In \emph{Advances in Neural Information Processing Systems}, volume~35, pages 30016--30030. Curran Associates, Inc.

\bibitem[{Ibrahim et~al.(2024)Ibrahim, Thérien, Gupta, Richter, Anthony, Lesort, Belilovsky, and Rish}]{ibrahim2024simple}
Adam Ibrahim, Benjamin Thérien, Kshitij Gupta, Mats~L. Richter, Quentin Anthony, Timothée Lesort, Eugene Belilovsky, and Irina Rish. 2024.
\newblock \href {http://arxiv.org/abs/2403.08763} {Simple and scalable strategies to continually pre-train large language models}.

\bibitem[{Joshi et~al.(2020)Joshi, Santy, Budhiraja, Bali, and Choudhury}]{joshi2020state}
Pratik Joshi, Sebastin Santy, Amar Budhiraja, Kalika Bali, and Monojit Choudhury. 2020.
\newblock \href {https://doi.org/10.18653/v1/2020.acl-main.560} {The state and fate of linguistic diversity and inclusion in the {NLP} world}.
\newblock In \emph{Proceedings of the 58th Annual Meeting of the Association for Computational Linguistics}, pages 6282--6293.

\bibitem[{Kew et~al.(2023)Kew, Schottmann, and Sennrich}]{kew2023turning}
Tannon Kew, Florian Schottmann, and Rico Sennrich. 2023.
\newblock \href {http://arxiv.org/abs/2312.12683} {Turning english-centric {LLMs} into polyglots: How much multilinguality is needed?}

\bibitem[{Kocetkov et~al.(2023)Kocetkov, Li, allal, LI, Mou, Jernite, Mitchell, Ferrandis, Hughes, Wolf, Bahdanau, Werra, and de~Vries}]{kocetkov2023stack}
Denis Kocetkov, Raymond Li, Loubna~Ben allal, Jia LI, Chenghao Mou, Yacine Jernite, Margaret Mitchell, Carlos~Mu{\~n}oz Ferrandis, Sean Hughes, Thomas Wolf, Dzmitry Bahdanau, Leandro~Von Werra, and Harm de~Vries. 2023.
\newblock \href {https://openreview.net/forum?id=pxpbTdUEpD} {The stack: 3 {TB} of permissively licensed source code}.
\newblock \emph{Transactions on Machine Learning Research}.

\bibitem[{Le~Scao et~al.(2022)Le~Scao, Fan, Akiki, Pavlick, Ili{\'c}, Hesslow, Castagn{\'e}, Luccioni, Yvon, Gall{\'e} et~al.}]{lescao2022bloom}
Teven Le~Scao, Angela Fan, Christopher Akiki, Ellie Pavlick, Suzana Ili{\'c}, Daniel Hesslow, Roman Castagn{\'e}, Alexandra~Sasha Luccioni, Fran{\c{c}}ois Yvon, Matthias Gall{\'e}, et~al. 2022.
\newblock {BLOOM}: A 176b-parameter open-access multilingual language model.

\bibitem[{Li et~al.(2023)Li, Allal, Zi, Muennighoff, Kocetkov, Mou, Marone, Akiki, Li, Chim, Liu, Zheltonozhskii, Zhuo, Wang, Dehaene, Davaadorj, Lamy-Poirier, Monteiro, Shliazhko, Gontier, Meade, Zebaze, Yee, Umapathi, Zhu, Lipkin, Oblokulov, Wang, Murthy, Stillerman, Patel, Abulkhanov, Zocca, Dey, Zhang, Fahmy, Bhattacharyya, Yu, Singh, Luccioni, Villegas, Kunakov, Zhdanov, Romero, Lee, Timor, Ding, Schlesinger, Schoelkopf, Ebert, Dao, Mishra, Gu, Robinson, Anderson, Dolan-Gavitt, Contractor, Reddy, Fried, Bahdanau, Jernite, Ferrandis, Hughes, Wolf, Guha, von Werra, and de~Vries}]{li2023starcoder}
Raymond Li, Loubna~Ben Allal, Yangtian Zi, Niklas Muennighoff, Denis Kocetkov, Chenghao Mou, Marc Marone, Christopher Akiki, Jia Li, Jenny Chim, Qian Liu, Evgenii Zheltonozhskii, Terry~Yue Zhuo, Thomas Wang, Olivier Dehaene, Mishig Davaadorj, Joel Lamy-Poirier, João Monteiro, Oleh Shliazhko, Nicolas Gontier, Nicholas Meade, Armel Zebaze, Ming-Ho Yee, Logesh~Kumar Umapathi, Jian Zhu, Benjamin Lipkin, Muhtasham Oblokulov, Zhiruo Wang, Rudra Murthy, Jason Stillerman, Siva~Sankalp Patel, Dmitry Abulkhanov, Marco Zocca, Manan Dey, Zhihan Zhang, Nour Fahmy, Urvashi Bhattacharyya, Wenhao Yu, Swayam Singh, Sasha Luccioni, Paulo Villegas, Maxim Kunakov, Fedor Zhdanov, Manuel Romero, Tony Lee, Nadav Timor, Jennifer Ding, Claire Schlesinger, Hailey Schoelkopf, Jan Ebert, Tri Dao, Mayank Mishra, Alex Gu, Jennifer Robinson, Carolyn~Jane Anderson, Brendan Dolan-Gavitt, Danish Contractor, Siva Reddy, Daniel Fried, Dzmitry Bahdanau, Yacine Jernite, Carlos~Muñoz Ferrandis, Sean Hughes, Thomas Wolf, Arjun Guha, Leandro von
  Werra, and Harm de~Vries. 2023.
\newblock \href {http://arxiv.org/abs/2305.06161} {Starcoder: may the source be with you!}

\bibitem[{Lin et~al.(2024)Lin, Ravichander, Lu, Dziri, Sclar, Chandu, Bhagavatula, and Choi}]{Lin2024ReAlign}
Bill~Yuchen Lin, Abhilasha Ravichander, Ximing Lu, Nouha Dziri, Melanie Sclar, Khyathi Chandu, Chandra Bhagavatula, and Yejin Choi. 2024.
\newblock \href {https://arxiv.org/abs/2312.01552} {The unlocking spell on base llms: Rethinking alignment via in-context learning}.
\newblock In \emph{International Conference on Learning Representations}.

\bibitem[{Lin et~al.(2022{\natexlab{a}})Lin, Hilton, and Evans}]{lin2022truthfulqa}
Stephanie Lin, Jacob Hilton, and Owain Evans. 2022{\natexlab{a}}.
\newblock Truthfulqa: Measuring how models mimic human falsehoods.
\newblock In \emph{Proceedings of the 60th Annual Meeting of the Association for Computational Linguistics (Volume 1: Long Papers)}, pages 3214--3252.

\bibitem[{Lin et~al.(2022{\natexlab{b}})Lin, Mihaylov, Artetxe, Wang, Chen, Simig, Ott, Goyal, Bhosale, Du, Pasunuru, Shleifer, Koura, Chaudhary, O'Horo, Wang, Zettlemoyer, Kozareva, Diab, Stoyanov, and Li}]{lin2022fewshot}
Xi~Victoria Lin, Todor Mihaylov, Mikel Artetxe, Tianlu Wang, Shuohui Chen, Daniel Simig, Myle Ott, Naman Goyal, Shruti Bhosale, Jingfei Du, Ramakanth Pasunuru, Sam Shleifer, Punit~Singh Koura, Vishrav Chaudhary, Brian O'Horo, Jeff Wang, Luke Zettlemoyer, Zornitsa Kozareva, Mona Diab, Veselin Stoyanov, and Xian Li. 2022{\natexlab{b}}.
\newblock \href {http://arxiv.org/abs/2112.10668} {Few-shot learning with multilingual language models}.

\bibitem[{Lozhkov et~al.(2024)Lozhkov, Li, Allal, Cassano, Lamy-Poirier, Tazi, Tang, Pykhtar, Liu, Wei, Liu, Tian, Kocetkov, Zucker, Belkada, Wang, Liu, Abulkhanov, Paul, Li, Li, Risdal, Li, Zhu, Zhuo, Zheltonozhskii, Dade, Yu, Krauß, Jain, Su, He, Dey, Abati, Chai, Muennighoff, Tang, Oblokulov, Akiki, Marone, Mou, Mishra, Gu, Hui, Dao, Zebaze, Dehaene, Patry, Xu, McAuley, Hu, Scholak, Paquet, Robinson, Anderson, Chapados, Patwary, Tajbakhsh, Jernite, Ferrandis, Zhang, Hughes, Wolf, Guha, von Werra, and de~Vries}]{lozhkov2024starcoder2}
Anton Lozhkov, Raymond Li, Loubna~Ben Allal, Federico Cassano, Joel Lamy-Poirier, Nouamane Tazi, Ao~Tang, Dmytro Pykhtar, Jiawei Liu, Yuxiang Wei, Tianyang Liu, Max Tian, Denis Kocetkov, Arthur Zucker, Younes Belkada, Zijian Wang, Qian Liu, Dmitry Abulkhanov, Indraneil Paul, Zhuang Li, Wen-Ding Li, Megan Risdal, Jia Li, Jian Zhu, Terry~Yue Zhuo, Evgenii Zheltonozhskii, Nii Osae~Osae Dade, Wenhao Yu, Lucas Krauß, Naman Jain, Yixuan Su, Xuanli He, Manan Dey, Edoardo Abati, Yekun Chai, Niklas Muennighoff, Xiangru Tang, Muhtasham Oblokulov, Christopher Akiki, Marc Marone, Chenghao Mou, Mayank Mishra, Alex Gu, Binyuan Hui, Tri Dao, Armel Zebaze, Olivier Dehaene, Nicolas Patry, Canwen Xu, Julian McAuley, Han Hu, Torsten Scholak, Sebastien Paquet, Jennifer Robinson, Carolyn~Jane Anderson, Nicolas Chapados, Mostofa Patwary, Nima Tajbakhsh, Yacine Jernite, Carlos~Muñoz Ferrandis, Lingming Zhang, Sean Hughes, Thomas Wolf, Arjun Guha, Leandro von Werra, and Harm de~Vries. 2024.
\newblock \href {http://arxiv.org/abs/2402.19173} {Starcoder 2 and the stack v2: The next generation}.

\bibitem[{Luukkonen et~al.(2023)Luukkonen, Komulainen, Luoma, Eskelinen, Kanerva, Kupari, Ginter, Laippala, Muennighoff, Piktus, Wang, Tazi, Scao, Wolf, Suominen, Sairanen, Merioksa, Heinonen, Vahtola, Antao, and Pyysalo}]{luukkonen2023fingpt}
Risto Luukkonen, Ville Komulainen, Jouni Luoma, Anni Eskelinen, Jenna Kanerva, Hanna-Mari Kupari, Filip Ginter, Veronika Laippala, Niklas Muennighoff, Aleksandra Piktus, Thomas Wang, Nouamane Tazi, Teven Scao, Thomas Wolf, Osma Suominen, Samuli Sairanen, Mikko Merioksa, Jyrki Heinonen, Aija Vahtola, Samuel Antao, and Sampo Pyysalo. 2023.
\newblock \href {https://doi.org/10.18653/v1/2023.emnlp-main.164} {{F}in{GPT}: Large generative models for a small language}.
\newblock In \emph{Proceedings of the 2023 Conference on Empirical Methods in Natural Language Processing}, pages 2710--2726. Association for Computational Linguistics.

\bibitem[{Madaan et~al.(2022)Madaan, Zhou, Alon, Yang, and Neubig}]{madaan2022language}
Aman Madaan, Shuyan Zhou, Uri Alon, Yiming Yang, and Graham Neubig. 2022.
\newblock Language models of code are few-shot commonsense learners.
\newblock In \emph{Proceedings of the 2022 Conference on Empirical Methods in Natural Language Processing}, pages 1384--1403.

\bibitem[{{MosaicML}(2023)}]{MosaicML2023Introducing}
{MosaicML}. 2023.
\newblock \href {https://www.databricks.com/blog/mpt-30b} {Introducing {MPT-30B}: Raising the bar for open-source foundation models}.

\bibitem[{Muennighoff et~al.(2024)Muennighoff, Rush, Barak, Le~Scao, Tazi, Piktus, Pyysalo, Wolf, and Raffel}]{muennighoff2024scaling}
Niklas Muennighoff, Alexander Rush, Boaz Barak, Teven Le~Scao, Nouamane Tazi, Aleksandra Piktus, Sampo Pyysalo, Thomas Wolf, and Colin~A Raffel. 2024.
\newblock Scaling data-constrained language models.
\newblock \emph{Advances in Neural Information Processing Systems}, 36.

\bibitem[{Petrov et~al.(2023)Petrov, La~Malfa, Torr, and Bibi}]{petrov2023language}
Aleksandar Petrov, Emanuele La~Malfa, Philip~HS Torr, and Adel Bibi. 2023.
\newblock Language model tokenizers introduce unfairness between languages.
\newblock \emph{arXiv preprint arXiv:2305.15425}.

\bibitem[{Pfeiffer et~al.(2022)Pfeiffer, Goyal, Lin, Li, Cross, Riedel, and Artetxe}]{pfeiffer2022lifting}
Jonas Pfeiffer, Naman Goyal, Xi~Lin, Xian Li, James Cross, Sebastian Riedel, and Mikel Artetxe. 2022.
\newblock \href {https://doi.org/10.18653/v1/2022.naacl-main.255} {Lifting the curse of multilinguality by pre-training modular transformers}.
\newblock In \emph{Proceedings of the 2022 Conference of the North American Chapter of the Association for Computational Linguistics: Human Language Technologies}, pages 3479--3495, Seattle, United States. Association for Computational Linguistics.

\bibitem[{Press et~al.(2021)Press, Smith, and Lewis}]{press2021train}
Ofir Press, Noah~A Smith, and Mike Lewis. 2021.
\newblock Train short, test long: Attention with linear biases enables input length extrapolation.
\newblock \emph{arXiv preprint arXiv:2108.12409}.

\bibitem[{Pyysalo et~al.(2021)Pyysalo, Kanerva, Virtanen, and Ginter}]{pyysalo2021wikibert}
Sampo Pyysalo, Jenna Kanerva, Antti Virtanen, and Filip Ginter. 2021.
\newblock {WikiBERT} models: Deep transfer learning for many languages.
\newblock \emph{NoDaLiDa 2021}, page~1.

\bibitem[{Radford et~al.(2018)Radford, Narasimhan, Salimans, and Sutskever}]{radford2018improving}
Alec Radford, Karthik Narasimhan, Tim Salimans, and Ilya Sutskever. 2018.
\newblock Improving language understanding by generative pre-training.

\bibitem[{Rust et~al.(2021)Rust, Pfeiffer, Ruder, and Gurevych}]{rust2021good}
Phillip Rust, Ivan Pfeiffer, Jonas an~Vuli{\'c}, Sebastian Ruder, and Iryna Gurevych. 2021.
\newblock \href {https://doi.org/10.18653/v1/2021.acl-long.243} {How good is your tokenizer? on the monolingual performance of multilingual language models}.
\newblock In \emph{Proceedings of the 59th Annual Meeting of the Association for Computational Linguistics and the 11th International Joint Conference on Natural Language Processing (Volume 1: Long Papers)}, pages 3118--3135.

\bibitem[{Sakaguchi et~al.(2019)Sakaguchi, Bras, Bhagavatula, and Choi}]{sakaguchi2019winogrande}
Keisuke Sakaguchi, Ronan~Le Bras, Chandra Bhagavatula, and Yejin Choi. 2019.
\newblock Winogrande: An adversarial winograd schema challenge at scale.
\newblock \emph{arXiv preprint arXiv:1907.10641}.

\bibitem[{Sardana and Frankle(2023)}]{sardana2023beyond}
Nikhil Sardana and Jonathan Frankle. 2023.
\newblock Beyond chinchilla-optimal: Accounting for inference in language model scaling laws.
\newblock \emph{arXiv preprint arXiv:2401.00448}.

\bibitem[{Soboleva et~al.(2023)Soboleva, Al-Khateeb, Myers, Steeves, Hestness, and Dey}]{cerebras2023slimpajama}
Daria Soboleva, Faisal Al-Khateeb, Robert Myers, Jacob~R Steeves, Joel Hestness, and Nolan Dey. 2023.
\newblock \href {https://huggingface.co/datasets/cerebras/SlimPajama-627B} {{SlimPajama: A 627B token cleaned and deduplicated version of RedPajama}}.

\bibitem[{Soldaini et~al.(2024)Soldaini, Kinney, Bhagia, Schwenk, Atkinson, Authur, Bogin, Chandu, Dumas, Elazar, Hofmann, Jha, Kumar, Lucy, Lyu, Lambert, Magnusson, Morrison, Muennighoff, Naik, Nam, Peters, Ravichander, Richardson, Shen, Strubell, Subramani, Tafjord, Walsh, Zettlemoyer, Smith, Hajishirzi, Beltagy, Groeneveld, Dodge, and Lo}]{dolma}
Luca Soldaini, Rodney Kinney, Akshita Bhagia, Dustin Schwenk, David Atkinson, Russell Authur, Ben Bogin, Khyathi Chandu, Jennifer Dumas, Yanai Elazar, Valentin Hofmann, Ananya~Harsh Jha, Sachin Kumar, Li~Lucy, Xinxi Lyu, Nathan Lambert, Ian Magnusson, Jacob Morrison, Niklas Muennighoff, Aakanksha Naik, Crystal Nam, Matthew~E. Peters, Abhilasha Ravichander, Kyle Richardson, Zejiang Shen, Emma Strubell, Nishant Subramani, Oyvind Tafjord, Pete Walsh, Luke Zettlemoyer, Noah~A. Smith, Hannaneh Hajishirzi, Iz~Beltagy, Dirk Groeneveld, Jesse Dodge, and Kyle Lo. 2024.
\newblock {Dolma: an Open Corpus of Three Trillion Tokens for Language Model Pretraining Research}.
\newblock \emph{arXiv preprint}.

\bibitem[{Srivastava et~al.(2022)Srivastava, Rastogi, Rao, Shoeb, Abid, Fisch, Brown, Santoro, Gupta, Garriga-Alonso et~al.}]{srivastava2022beyond}
Aarohi Srivastava, Abhinav Rastogi, Abhishek Rao, Abu Awal~Md Shoeb, Abubakar Abid, Adam Fisch, Adam~R Brown, Adam Santoro, Aditya Gupta, Adri{\`a} Garriga-Alonso, et~al. 2022.
\newblock Beyond the imitation game: Quantifying and extrapolating the capabilities of language models.
\newblock \emph{arXiv preprint arXiv:2206.04615}.

\bibitem[{Tiedemann(2009)}]{tiedemann2009news}
J{\"o}rg Tiedemann. 2009.
\newblock News from {OPUS}-a collection of multilingual parallel corpora with tools and interfaces.
\newblock In \emph{Recent advances in natural language processing}, volume~5, pages 237--248.

\bibitem[{Tiedemann(2020)}]{tiedemann2020tatoeba}
J{\"o}rg Tiedemann. 2020.
\newblock \href {https://www.aclweb.org/anthology/2020.wmt-1.139} {The {T}atoeba {T}ranslation {C}hallenge {--} {R}ealistic data sets for low resource and multilingual {MT}}.
\newblock In \emph{Proceedings of the Fifth Conference on Machine Translation}, pages 1174--1182, Online. Association for Computational Linguistics.

\bibitem[{Tiedemann and Thottingal(2020)}]{TiedemannThottingal:EAMT2020}
J{\"o}rg Tiedemann and Santhosh Thottingal. 2020.
\newblock {OPUS-MT} — {B}uilding open translation services for the {W}orld.
\newblock In \emph{Proceedings of the 22nd Annual Conferenec of the European Association for Machine Translation (EAMT)}, Lisbon, Portugal.

\bibitem[{{Together Computer}(2023)}]{together2023redpajama}
{Together Computer}. 2023.
\newblock \href {https://github.com/togethercomputer/RedPajama-Data} {Redpajama: An open source recipe to reproduce llama training dataset}.

\bibitem[{Touvron et~al.(2023)Touvron, Lavril, Izacard, Martinet, Lachaux, Lacroix, Rozière, Goyal, Hambro, Azhar, Rodriguez, Joulin, Grave, and Lample}]{touvron2023llama}
Hugo Touvron, Thibaut Lavril, Gautier Izacard, Xavier Martinet, Marie-Anne Lachaux, Timothée Lacroix, Baptiste Rozière, Naman Goyal, Eric Hambro, Faisal Azhar, Aurelien Rodriguez, Armand Joulin, Edouard Grave, and Guillaume Lample. 2023.
\newblock \href {http://arxiv.org/abs/2302.13971} {Llama: Open and efficient foundation language models}.

\bibitem[{Ubierna et~al.(2022)Ubierna, Santos, and Mercier-Blais}]{ubierna2022water}
Mar{\'\i}a Ubierna, Cristina~D{\'\i}ez Santos, and Sara Mercier-Blais. 2022.
\newblock Water security and climate change: hydropower reservoir greenhouse gas emissions.
\newblock \emph{Water Security Under Climate Change}, pages 69--94.

\bibitem[{Vaswani et~al.(2017)Vaswani, Shazeer, Parmar, Uszkoreit, Jones, Gomez, Kaiser, and Polosukhin}]{vaswani2017attention}
Ashish Vaswani, Noam Shazeer, Niki Parmar, Jakob Uszkoreit, Llion Jones, Aidan~N Gomez, {\L}ukasz Kaiser, and Illia Polosukhin. 2017.
\newblock Attention is all you need.
\newblock \emph{Advances in neural information processing systems}, 30.

\bibitem[{Vilar et~al.(2023)Vilar, Freitag, Cherry, Luo, Ratnakar, and Foster}]{vilar2023prompting}
David Vilar, Markus Freitag, Colin Cherry, Jiaming Luo, Viresh Ratnakar, and George Foster. 2023.
\newblock Prompting palm for translation: Assessing strategies and performance.
\newblock In \emph{Proceedings of the 61st Annual Meeting of the Association for Computational Linguistics (Volume 1: Long Papers)}, pages 15406--15427.

\bibitem[{Villalobos et~al.(2022)Villalobos, Sevilla, Heim, Besiroglu, Hobbhahn, and Ho}]{villalobos2022will}
Pablo Villalobos, Jaime Sevilla, Lennart Heim, Tamay Besiroglu, Marius Hobbhahn, and Anson Ho. 2022.
\newblock Will we run out of data? an analysis of the limits of scaling datasets in machine learning.
\newblock \emph{arXiv preprint arXiv:2211.04325}.

\bibitem[{Wei et~al.(2023)Wei, Wei, Lin, Li, Zhang, Ren, Li, Wan, Cao, Xie et~al.}]{wei2023polylm}
Xiangpeng Wei, Haoran Wei, Huan Lin, Tianhao Li, Pei Zhang, Xingzhang Ren, Mei Li, Yu~Wan, Zhiwei Cao, Binbin Xie, et~al. 2023.
\newblock {PolyLM}: An open source polyglot large language model.
\newblock \emph{arXiv preprint arXiv:2307.06018}.

\bibitem[{Zellers et~al.(2019)Zellers, Holtzman, Bisk, Farhadi, and Choi}]{zellers2019hellaswag}
Rowan Zellers, Ari Holtzman, Yonatan Bisk, Ali Farhadi, and Yejin Choi. 2019.
\newblock \href {https://doi.org/10.18653/v1/P19-1472} {{H}ella{S}wag: Can a machine really finish your sentence?}
\newblock In \emph{Proceedings of the 57th Annual Meeting of the Association for Computational Linguistics}, pages 4791--4800. Association for Computational Linguistics.

\bibitem[{Zhao et~al.(2024)Zhao, Zhang, Gao, Zhang, Gui, and Huang}]{zhao2024llama}
Jun Zhao, Zhihao Zhang, Luhui Gao, Qi~Zhang, Tao Gui, and Xuanjing Huang. 2024.
\newblock \href {http://arxiv.org/abs/2401.01055} {{LLaMA} beyond english: An empirical study on language capability transfer}.

\bibitem[{Zheng et~al.(2023)Zheng, Chiang, Sheng, Zhuang, Wu, Zhuang, Lin, Li, Li, Xing, Zhang, Gonzalez, and Stoica}]{zheng2023judging}
Lianmin Zheng, Wei-Lin Chiang, Ying Sheng, Siyuan Zhuang, Zhanghao Wu, Yonghao Zhuang, Zi~Lin, Zhuohan Li, Dacheng Li, Eric~P. Xing, Hao Zhang, Joseph~E. Gonzalez, and Ion Stoica. 2023.
\newblock \href {http://arxiv.org/abs/2306.05685} {{Judging LLM-as-a-Judge with MT-Bench and Chatbot Arena}}.

\bibitem[{Zhu et~al.(2023)Zhu, Liu, Dong, Xu, Huang, Kong, Chen, and Li}]{zhu2023multilingual}
Wenhao Zhu, Hongyi Liu, Qingxiu Dong, Jingjing Xu, Shujian Huang, Lingpeng Kong, Jiajun Chen, and Lei Li. 2023.
\newblock \href {http://arxiv.org/abs/2304.04675} {Multilingual machine translation with large language models: Empirical results and analysis}.

\end{thebibliography}

\appendix
\clearpage
\onecolumn
\section{Appendix}
\label{sec:appendix}

\subsection{Training details}
\label{sec:training_details}

It has been our aim throughout this work to release Poro 34B fully openly, including model weights, pretraining configuration, the pretraining and evaluation data, and all associated scripts and tools. We provide here additional details of these to facilitate accurate reproduction of our work. The pretraining data sources are detailed in Table~\ref{tab:data_sources}, and the model and pretraining hyperparameters in Table~\ref{tab:model_arch_params}. 

\begin{table*}[h!]
\centering
\setlength{\tabcolsep}{3pt}
\begin{tabular}{lll}
\toprule
Dataset    & Language & Reference \\ 
\midrule
SlimPajama & English  & \url{https://huggingface.co/datasets/cerebras/SlimPajama-627B} \\
Starcoder  & Code     & \url{https://huggingface.co/datasets/bigcode/starcoderdata} \\
Tatoeba challenge & Eng-Fin & \url{https://huggingface.co/datasets/tatoeba} \\
Project Gutenberg      & English  & \url{https://huggingface.co/datasets/allenai/dolma} \\
Parsebank   &  Finnish & \url{https://turkunlp.org/finnish_nlp.html} \\
mC4         &\ditto & \url{https://huggingface.co/datasets/mc4} \\
CC-Fi       &\ditto         & \url{https://github.com/TurkuNLP/CC-Fi} \\
Fiwiki      &\ditto         & \url{https://fi.wikipedia.org/wiki} \\
Lönnrot     &\ditto         & \url{http://www.lonnrot.net} \\
Suomi24     &\ditto & \url{http://urn.fi/urn:nbn:fi:lb-2021101527} \\
Reddit-Fi   &\ditto & \url{https://www.reddit.com/r/Suomi} \\
STT         &\ditto & \url{http://urn.fi/urn:nbn:fi:lb-2019041501} \\
Yle         &\ditto & \url{http://urn.fi/urn:nbn:fi:lb-2017070501} \\
Yle         &\ditto & \url{http://urn.fi/urn:nbn:fi:lb-2021050401} \\
Yle         &\ditto  & \url{http://urn.fi/urn:nbn:fi:lb-2019050901} \\
Yle         &\ditto  & \url{http://urn.fi/urn:nbn:fi:lb-2021050701} \\
\bottomrule
\end{tabular}
\caption{Data sources}
\label{tab:data_sources}
\end{table*}

\begin{table*}[h!]
\centering
\begin{tabular}{lc|lc}
\toprule
\textit{Architecture hyperparameters} & & \textit{Pretraining hyperparameters} & \\
\midrule
Parameters & 34B & Global Batch Size & 2048 \\
Precision & bfloat16 & Learning rate & 1.5e-4 \\
Layers & 54 & Total tokens & 1000B \\
Hidden dim & 7168 & Warmup tokens & 10B \\
Attention heads & 56 & Decay tokens & 1000B \\
Vocab size & 131072 & Decay style & cosine \\
Sequence length & 2048 & Min. learning rate & 2e-5 \\
Activation & GELU & Adam ($\beta_1$, $\beta_2$) & (0.9, 0.95) \\
Position embedding & ALiBi & Weight decay & 2e-5 \\
Tied embeddings & True & Gradient clipping & 1.0 \\
\bottomrule
\end{tabular}
\caption{Model and training hyperparameters}
\label{tab:model_arch_params}
\end{table*}

\clearpage
\subsection{Detailed benchmark results}
\label{sec:benchmark_details}

Tables~\ref{tab:finnish-results},~\ref{tab:english-results}, and~\ref{tab:code-results} show the detailed benchmark results for Finnish, English, and code.

\begin{table*}[ht!]
\centering
\setlength{\tabcolsep}{2pt}
\begin{tabular}{l|c c c c|c c|c}
\bf Benchmark & \bf Poro~34B & \bf Llama 33B & \bf MPT-30b & \bf Falcon-40b & \bf FinGPT 8B & \bf FinGPT 13B & \bf Starcoder \\
\hline
Analogies & \bf 77.69 & 61.54 & 57.69 & 43.85 & 40.0 & 36.15 & 46.15 \\
Arithmetic & 54.28 & 47.74 & \bf 57.25 & 51.06 & 41.96 & 45.23 & 48.41 \\
Cause and Effect & 67.97 & 60.78 & 58.82 & 46.41 & 66.01 & \bf 69.28 & 54.90 \\
Emotions & \bf 55.00 & 45.00 & 39.37 & 16.88 & 45.62 & 38.75 & 23.13 \\
Empirical Judg. & \bf 62.63 & 43.43 & 43.43 & 34.34 & 32.32 & 36.36 & 44.44 \\
General Knowl. & \bf 75.71 & 48.57 & 37.14 & 22.86 & 51.43 & 40.00 & 22.86 \\
Intent Recogn. & \bf 83.24 & 77.75 & 77.31 & 46.24 & 51.43 & 58.24 & 65.03 \\
Misconceptions & \bf 53.73 & 51.49 & 50.00 & 50.00 & 51.45 & 45.52 & 47.01 \\
Paraphrase & \bf 58.50 & 53.00 & 52.50 & 54.50 & 49.50 & 45.50 & 47.50 \\
Sentence Ambig. & \bf 66.67 & 45.00 & 56.67 & 48.33 & 48.33 & 53.33 & 51.67 \\
Similarities Abst. & \bf 73.68 & 52.63 & 55.26 & 53.95 & 68.42 & 69.74 & 50.00 \\
\hline
Average & \bf 66.28 & 53.36 & 53.22 & 42.58 & 49.69 & 48.92 & 45.55 \\
\end{tabular}
\caption{FIN-Bench Finnish benchmark results}
\label{tab:finnish-results}
\end{table*}

\begin{table*}[ht!]
\centering
\setlength{\tabcolsep}{2pt}
\begin{tabular}{l|c c c c|c c|c}
\bf Benchmark & \bf Poro~34B & \bf Llama 33B & \bf MPT-30b & \bf Falcon-40b  & \bf FinGPT 8B & \bf FinGPT 13B & \bf Starcoder \\
\hline
ARC-Challenge & 53.16 & \bf 61.61 & 55.80 & 50.51 & 25.34 & 24.31 & 30.29 \\
Hellaswag & 77.77 & \bf 84.64 & 82.23 & 77.01 & 42.91 & 46.77 & 47.22 \\
MMLU & 46.29 & \bf 58.13 & 47.27 & 46.13 & 23.34 & 23.64 & 32.11 \\
TruthfulQA & 41.66 & 42.84 & 38.44 & 41.64 & 43.80 & \bf 44.58 & 40.06 \\
Winogrande & 72.77 & 80.27 & 74.82 & \bf 81.53 & 53.19 & 57.53 & 54.85 \\
GSM8K & 11.75 & \bf 32.27 & 17.13 & 2.43 & 0.22 & 0.22 & 8.11 \\
\hline
Average & 50.57 & \bf 59.96 & 52.62 & 49.87 & 31.47 & 32.85 & 35.44 \\
\end{tabular}
\caption{English benchmark results}
\label{tab:english-results}
\end{table*}

\begin{table*}[ht!]
\centering
\begin{tabular}{l|l|c c c c|c}
\bf Benchmark & \bf Category & \bf Poro~34B & \bf Llama 33B & \bf MPT-30b & \bf Falcon-40b & \bf Starcoder \\
\hline
HumanEval & Python & 37.20 & 34.15 & 35.37 & 34.15 & \bf 45.12 \\
MBPP & Python & 47.40 & 41.20 & 43.00 & 43.00 & \bf 53.00 \\
\hline 
 Average & & 41.80 & 37.67 & 39.18 & 38.57 & \bf 49.06\\
\end{tabular}
\caption{Code benchmark results}
\label{tab:code-results}
\end{table*}

\subsection{Hardware}
\label{sec:appendix_hardware}

Poro 34B was trained on the LUMI-G GPU partition of the LUMI supercomputer, located in Finland. LUMI is, at the time of this writing, the third fastest supercomputer in Europe, and the 8th fastest in the world (\url{https://www.top500.org/}). LUMI is also ranked 7th greenest by the Green500 list (\url{https://www.top500.org/lists/green500/}).  

The LUMI-G partition has 2978 nodes, with each node having four AMD MI250x GPUs with 128GB of memory each, and a single 64-core CPU. The MI250x is a multi-chip module (MCM), with dual-GCD (graphics compute die) design, which in practice means a node has eight logical devices, each logical device with access to 64GB of high bandwidth memory.

Each node has four 200Gbps Slingshot-11 network interconnects.  The nodes are connected together in a dragonfly topology. During benchmarking and scale testing we did not observe the network topology as a limiting factor for the required collective operation sizes. The total of 800 Gbps per-node bandwidth proved to be more than sufficient, and the communication overhead was minimal during training. 

\end{document}